\documentclass[format=acmsmall, review=false, screen=true]{acmart}

\usepackage{natbib}
\usepackage[T1]{fontenc}
\usepackage{float}
\usepackage{booktabs} 
\usepackage{subcaption}
\usepackage[ruled]{algorithm2e} 

\SetAlFnt{\small}
\SetAlCapFnt{\small}
\SetAlCapNameFnt{\small}
\SetAlCapHSkip{0pt}
\IncMargin{-\parindent}
\usepackage{footnote}
\usepackage{longtable}
\acmYear{2019}

\acmDOI{0000001.0000001}

\begin{document}
\title[Video Object Segmentation and Tracking: A Survey]{Video Object Segmentation and Tracking: A Survey}  
\author{Rui Yao}
\affiliation{%
  \institution{China University of Mining and Technology}
  \department{School of Computer Science and Engineering}
  \streetaddress{1 Daxue Rd}
  \city{Xuzhou}
  \state{Jingsu}
  \postcode{221116}
  \country{China}
 }
\author{Guosheng Lin}
\affiliation{%
  \institution{Nanyang Technological University}
  \department{School of Computer Science and Engineering}
  \country{Singapore}
}
\author{Shixiong Xia}
\author{Jiaqi Zhao}
\author{Yong Zhou}
\affiliation{%
  \institution{China University of Mining and Technology}
  \department{School of Computer Science and Engineering}
  \streetaddress{1 Daxue Rd}
  \city{Xuzhou}
  \state{Jingsu}
  \postcode{221116}
  \country{China}
}

\begin{abstract}
Object segmentation and object tracking are fundamental research area in the computer vision community. 
These two topics are difficult to handle some common challenges, such as occlusion, deformation, motion blur, and scale variation. The former contains heterogeneous object, interacting object, edge ambiguity, and shape complexity. And the latter suffers from difficulties in handling fast motion, out-of-view, and real-time processing. 
Combining the two problems of video object segmentation and tracking (VOST) can overcome their respective difficulties and improve their performance. 
VOST can be widely applied to many practical applications such as video summarization, high definition video compression, human computer interaction, and autonomous vehicles. 
This article aims to provide a comprehensive review of the state-of-the-art tracking methods, and classify these methods into different categories, and identify new trends.
First, we provide a hierarchical categorization existing approaches, including unsupervised VOS, semi-supervised VOS, interactive VOS, weakly supervised VOS, and segmentation-based tracking methods.
Second, we provide a detailed discussion and overview of the technical characteristics of the different methods. 
Third, we summarize the characteristics of the related video dataset, and provide a variety of evaluation metrics.
Finally, we point out a set of interesting future works and draw our own conclusions.
\end{abstract}

%
%


%
%



\thanks{

Author's addresses: R. Yao, S. Xia, 
J. Zhao, and Y. Zhou, School of Computer Science and Technology, China University of Mining and Technology, Xuzhou, 221116, China; emails: \{ruiyao, xiasx, jiaqizhao, yzhou\}cumt.edu.cn; \ G. Lin, School of Computer Science and Engineering, Nanyang Technological University; email: gslin@ntu.edu.sg.}

\maketitle

\renewcommand{\shortauthors}{R. Yao et al.}

\section{Introduction}
\label{sec:intro}

The rapid development of intelligent mobile terminals and the Internet has led to an exponential increase in video data. In order to effectively analyze and use video big data, it is very urgent to automatically segment and track the objects of interest in the video. Video object segmentation and tracking are two basic tasks in field of computer vision. Object segmentation divides the pixels in the video frame into two subsets of the foreground target and the background region, and generates the object segmentation mask, which is the core problem of behavior recognition and video retrieval. Object tracking is used to determine the exact location of the target in the video image and generate the object bounding box, which is a necessary step for intelligent monitoring, big data video analysis and so on.

The segmentation and tracking problems of video objects seem to be independent, but they are actually inseparable. That is to say, the solution to one of the problems usually involves solving another problem implicitly or explicitly. Obviously, by solving the object segmentation problem, it is easy to get a solution to the object tracking problem. On the one hand, accurate segmentation results provide reliable object observations for tracking, which can solve problems such as occlusion, deformation, scaling, \emph{etc.}, and fundamentally avoid tracking failures. Although not so obvious, the same is true for object tracking problems, which must provide at least a coarse solution to the problem of object segmentation. On the other hand, accurate object tracking results can also guide the segmentation algorithm to determine the object position, which reduces the impact of object fast movement, complex background, similar objects, \emph{etc.}, and improves object segmentation performance. A lot of research work has noticed that the simultaneous processing of the object segmentation and tracking problems, which can overcome their respective difficulties and improve their performance. The related problems can be divided into two major tasks: video object segmentation (VOS) and video object tracking (VOT).

The goal of video object segmentation is to segment a particular object instance in the entire video sequence of the object mask on a manual or automatic first frame, causing great concern in the computer vision community. Recent VOS algorithms can be organized by their annotations. The unsupervised and interactive VOS methods denote the two extremes of the degree of user interaction with the method: at one extreme, the former can produce a coherent space-time region through the bottom-up process without any user input, that is, without any video-specific tags~\cite{irani1998unified,grundmann2010efficient,Brox2010,lee2011key,faktor2014video,li2018instance}. In contrast, the latter uses a strongly supervised interaction method that requires pixel-level precise segmentation of the first frame (human provisioning is very time consuming), but also the human needs to loop error correction system~\cite{li2005video,wang2014touchcut,benard2017interactive,caelles20182018,maninis2018deep}. There are semi-supervised VOS approaches between the two extremes, which requires manual annotation to define what is the foreground object and then automatically segment to the rest frames of the sequence~\cite{4270202,Tsai2012,Jain243,8100048,perazzi2017learning}. In addition, because of the convenience of collecting video-level labels, another way to supervise VOS is to produce masks of objects given the ~\cite{zhang2015semantic,tang2013discriminative} or natural language expressions~\cite{khoreva2018video}. However, as mentioned above, the VOS algorithm implicitly handles the process of tracking. That is, the bottom-up approach uses a spatio-temporal motion and appearance similarity to segment the video in a fully automated manner. These methods read multiple or all image frames at once to take full advantage of the context of multiple frames, and segment the precise object mask. The datasets evaluated by these methods are dominated by short-term videos. Moreover, because these methods iteratively optimize energy functions or fine-turns a deep network, so it can be slow.

In contrast to VOS, given a sequence of input images, the video object tracking method utilizes a class-specific detector to robustly predict the motion state (location, size, or orientation, \emph{etc.}) of the object in each frame. In general, most of VOT methods are especially suitable for processing long-term sequences. Since these methods only need to output the location, orientation or size of the object, the VOT method uses the online manner for fast processing. For example, tracking-by-detection methods utilize generative~\cite{ross2008incremental} and/or discriminative~\cite{hare2016struck,yao2012robust} appearance models to accurate estimate object state. The impressive results of these methods prove accurate and fast tracking.
However, most algorithms are limited to generating bounding boxes or ellipses for their output, so that when non-rigid and articulated motions are involved in the object, they are often subject to visual drift problems. To address this problem, part-based tracking methods~\cite{yao2017part,yao2017real} have been presented, but they still use part of the bounding box for object localization. In order to leverage the precision object masks and fast object location, segmentation-based tracking methods have been developed which combine video object segmentation and tracking~\cite{Bibby2008,Aeschliman2010,wen2015jots,yeo2017superpixel,wang2018fast}. Most of methods estimate the object results (\emph{i.e.} bounding boxes of the object or/and object masks) by a combination of bottom-up and top-down algorithms. The contours of deformable objects or articulated motions can be propagated using these methods efficiently.

In the past decade, a large number of video object segmentation and tracking (VOST) studies have been published in the literature. The field of VOST has a wide range of practical applications, including video summarization, high definition (HD) video compression, gesture control and human interaction. For instance, VOST methods are widely applied to video summarization that exploits visual object across multiple videos~\cite{chu2015video}, and provide a useful tool that assists video retrieval or web browsing~\cite{rochan2018video}. In the filed of video compression, VOST is used in video-coding standards MPEG-4 to implement content-based features and high coding efficiency~\cite{988659}. In particular, the VOST can encode the video shot as a still background mosaic obtained after compensating the moving object by utilizing the content-based representation provided by MPEG-4~\cite{colombari2007segmentation}. Moreover, VOST can estimate the non-rigid target to achieve accurate tracking positioning and mask description, which can identify its motion instructions~\cite{Wang2017A}. They can replace simple human body language, especially various gesture controls.


\subsection{Challenges and issues}
\label{sec:challenges}

Many problems in video object segmentation and tracking are very challenging. In general, VOS and VOT have some common challenges, such as background clutter, low resolution, occlusion, deformation, motion blur, scale variation, \emph{etc.} But there are some specific characteristics determined by the objectives and tasks, for example, objects in the VOT can be
complex due to fast motion, out-of-view, and real-time processing. In addition, segmenting and tracking the effects of heterogeneous object, interacting object, edge ambiguity, shape complexity, \emph{etc.} A more detailed description is given in~\cite{wu2013online,perazzi2016benchmark}.

\begin{figure}[t]
\centering
\includegraphics[width=.65\textwidth]{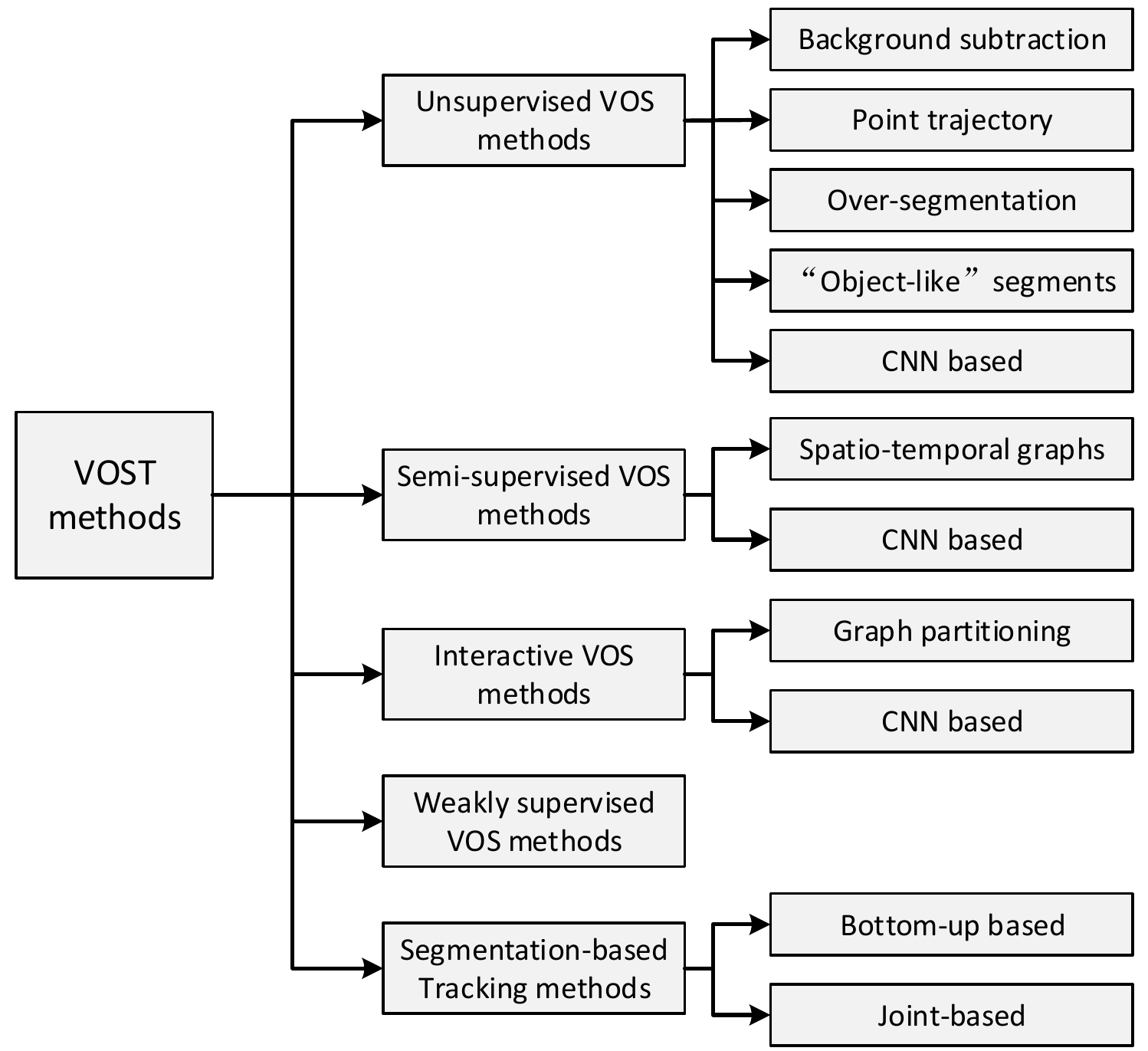}
\caption{Taxonomy of video object segmentation and tracking. }
\label{fig:overview_chart}
\end{figure}

To address these problems, tremendous progress has been made in the development of video object segmentation and tracking algorithms. These are mainly different from each other based on how they handle the following issues in visual segmentation and tracking: (i) which application scenario is suitable for VOST? (ii) Which object representation (\emph{i.e.} point, superpixel, patch, and object) is adapted to VOS? (iii) Which image features are appropriate for VOST? (iv) How to model the motion of an object in VOST? (v) How to per-process and post-process CNN-based VOS methods? (vi) Which datasets are suitable for the evaluation VOST, and what are their characteristics? A number of VOST methods have been proposed that attempt to answer these issues for various scenarios. Motivated by the objective, this survey divides the video object segmentation and tracking methods into broad categories and provides a comprehensive review of some representative approaches. We hope to help readers gain valuable VOST knowledge and choose the most appropriate application for their specific VOST tasks. In addition, we will discuss video object segmentation and tracking new trends in the community, and hope to provide several interesting ideas to new methods.

\subsection{Organization and contributions of this survey}
\label{sec:contributions}

As shown in Fig.~\ref{fig:overview_chart}, we summarize our organization in this survey. To investigate a suitable application scenario for VOST, we group these methods into five main categories: unsupervised VOS, semi-supervised VOS, interactive VOS, weakly supervised VOS, and segmentation-based tracking methods. 

The unsupervised VOS algorithm typically relies on certain restrictive assumptions about the application scenario, so it does not have to be manually annotated in the first frame. According to discover primary objects using appearance and motion cues, in Sec.~\ref{sec:unsupervised}, we categorize them as background subtraction, point trajectory, over-segmentation, ``object-like'' segments, and convolutional neural networks based methods. In Tab.~\ref{tab:unsupervised1}, we also summarize some object representation, for example, pixel, superpixel, supervoxel, and patch, and image features. In Sec.~\ref{sec:semi}, we describe the semi-supervised VOS methods for modeling the appearance representations and temporal connections, and performing segmentation and tracking jointly. In Tab.~\ref{tab:cnn_vos}, we discuss various of per-process and post-process CNN-based VOS methods. In Sec.~\ref{sec:interactive}, interactive VOS methods are summarized by the way of user interaction and motion cues. In Sec.~\ref{sec:weakly}, we discuss various weakly supervised information for video object segmentation. In Sec.~\ref{sec:track}, we group and describe the segmentation-based tracking methods, and explain the advantages or disadvantages of different bottom-up and joint-based frameworks, as shown in Tab.~\ref{tab:bottom-up} and Tab.~\ref{tab:joint-based}. In addition, we investigate a number of video datasets for video object segmentation and tracking, and explain the metrics of pixel-wise mask and bounding box based techniques. Finally, we present several interesting issues for the future research in Sec.~\ref{sec:future}, and help researchers in other related fields to explore the possible benefits of VOST techniques.

Although there are surveys on VOS~\cite{perazzi2016benchmark,Erdem2004} and VOT~\cite{yilmaz2006object,li2013survey,Wu2015Object}, they are not directly applicable to joint video object segmentation and tracking, unlike our surveys. First, Perazzi~\emph{et al.}~\cite{perazzi2016benchmark} present a dataset and evaluation metrics for VOS methods, Erdem~\emph{et al.}~\cite{Erdem2004} measure to evaluate quantitatively the performance of VOST methods in 2004. In comparison, we focuses on the summary of methods of video object segmentation, but also object tracking. Second, Yilmaz~\emph{et al.}~\cite{yilmaz2006object} and Li~\emph{et al.}~\cite{li2013survey} discuss generic object tracking algorithms, and Wu~\emph{et la.}~\cite{Wu2015Object} evaluate the performance of single object tracking, therefore, they are different from our segmentation-based tracking discussion.

In this survey, we provide a comprehensive review of video object segmentation and tracking, and summarize our contributions as follows: (i) As shown in Fig.~\ref{fig:overview_chart}, a hierarchical categorization existing approaches is provided in video object segmentation and tracking. We roughly classify methods into five categories. Then, for each category, different methods are further categorized. (ii) We provide a detailed discussion and overview of the technical characteristics of the different methods in unsupervised VOS, semi-supervised VOS, interactive VOS, and segmentation-based tracking. (iii) We summarize the characteristics of the related video dataset, and provide a variety of evaluation metrics.

\section{Major methods}
\label{methods}

In the section, video object segmentation and tracking methods are grouped into five categories: unsupervised video object segmentation methods, semi-supervised video object segmentation methods, interactive video object segmentation methods, weakly supervised video object segmentation methods, and segmentation-based tracking methods.

\subsection{Unsupervised video object segmentation}
\label{sec:unsupervised}

The unsupervised VOS algorithm does not require any user input, it can automatically find objects. In general, they assume that the objects to be segmented and tracked have different motions or appear frequently in the sequence of images. Following we will review and discuss five groups of the unsupervised methods.

\subsubsection{Background subtraction}
\label{sec:background}

Early video segmentation methods were primarily geometric based and limited to specific motion backgrounds. The classic background subtraction method simulates the background appearance of each pixel and treats rapidly changing pixels as foreground. Any significant change in the image and background model represents a moving object. The pixels that make up the changed region are marked for further processing. A connected component algorithm is used to estimate the connected region corresponding to the object. Therefore, the above process is called background subtraction. Video object segmentation is achieved by constructing a representation of the scene called the background model and then finding deviations from the model for each input frame. 

According to the dimension of the used motion, background subtraction methods can be divided into stationary backgrounds~\cite{stauffer2000learning,elgammal2002background,han2012density}, backgrounds undergoing 2D parametric motion~\cite{irani1994computing,ren2003statistical,criminisi2006bilayer,barnich2011vibe}, and backgrounding undergoing 3D motions~\cite{torr1998concerning,irani1998unified,brutzer2011evaluation}. 

\paragraph{Stationary backgrounds}
Background subtraction became popular following the work of Wren \emph{et al.}~\cite{wren1997pfinder}. They use a multiclass statistical model of color pixel, $I(x,y)$, of a stationary background with a single 3D ($Y, U$, and $V$ color space) Gaussian, $I(x,y)\sim N(\mu(x,y),\Sigma(x,y))$. The model parameters (the mean $\mu(x,y)$ and the covariance $\Sigma(x,y)$) are learned from the color observations in several consecutive frames. For each pixel $(x, y)$ in the input video frame, after the model of the background is derived, they calculate the likelihood that their color is from $N(\mu(x,y),\Sigma(x,y))$, and the deviation from the pixel. The foreground model is marked as a foreground pixel. However, Gao \emph{et al.}~\cite{gao2000error} show that a single Gaussian would be insufficient to model the pixel value while accounting for acquisition noise. Therefore, some work begin to improve the performance of background modeling by using a multimodal statistical model to describe the background color per pixel. For example, Stauffer and Grimson~\cite{stauffer2000learning} build models each pixel as a mixture of Gaussians (MoG) and uses an on-line approximation to update the model.  Rather than explicitly modeling the values of all the pixels as one particular type of distribution, they model the values of a particular pixel as a mixture of Gaussians. In~\cite{elgammal2002background}, Elgammal and Davis use nonparametric kernel density estimation (KDE) to model the per-pixel background. They construct a statistical representation of the scene background that supports sensitive detection of moving objects in the scene. In~\cite{han2012density}, the authors propose a pixelwise background modeling and subtraction technique using multiple features, where generative (kernel density approximation (KDA)) and discriminative (support vector machine (SVM)) techniques are combined for classification. 

\paragraph{Backgrounds undergoing 2D parametric motion}
Instead of modeling stationary backgrounds, another methods use backgrounds undergoing 2D parametric motion. For instance, Irani~\emph{et al.}~\cite{irani1994computing} detect and track occluding and transparent moving objects, and use temporal integration without assuming motion constancy. The temporal integration maintains sharpness of the tracked object, while blurring objects that have other motions. Ren \emph{et al.}~\cite{ren2003statistical} propose a background subtraction method based spatial distribution of Gaussians model for the foreground detection from a non-stationary background.
Criminisi \emph{et al.}~\cite{criminisi2006bilayer} present segmentation of videos by probabilistic fusion of motion, color and contrast cues together with spatial and temporal priors. They build the automatic layer separation and background substitution method. Barnich and Droogenbroeck~\cite{barnich2011vibe} introduce a universal sample-based background subtraction algorithm, which include pixel model and classification process, background model initialization, and updating the background model over time.

\paragraph{Backgrounding undergoing 3D motions}
Irani and Anandan~\cite{irani1998unified} describe a unified approach to handling moving-object detection in both 2D and 3D scenes. A two-dimensional algorithm applied when a scene can be approximated by a plane and when the camera is only rotated and scaled. 3D algorithm that works only when there is a significant depth change in the scene and the camera is translating. This method bridges the two extremes of the strategy of the gap.
Torr and Zisserman~\cite{torr1998concerning} present a Bayesian methods of motion segmentation using the constraints enforced by rigid motion. Each motion model build 3D relations and 2D relations. In~\cite{brutzer2011evaluation}, Brutzer~\emph{et al.} evaluate the background subtraction method.  They identify the main challenges of background subtraction, and then compare the performance of several background subtraction methods with post-processing. 

\emph{Discussion.} Due to the use of stationary backgrounds and 3D motions, these methods have different properties. Overall, the aforementioned methods must rely on the restrictive assumption that the camera is stable and slowly moving. That is, it is sensitive to model selection (2D or 3D) and cannot handle special backgrounds such as non-rigid object.

\subsubsection{Point trajectory}
\label{sec:point}

\begin{figure}[t]
\centering
\includegraphics[width=.16\textwidth]{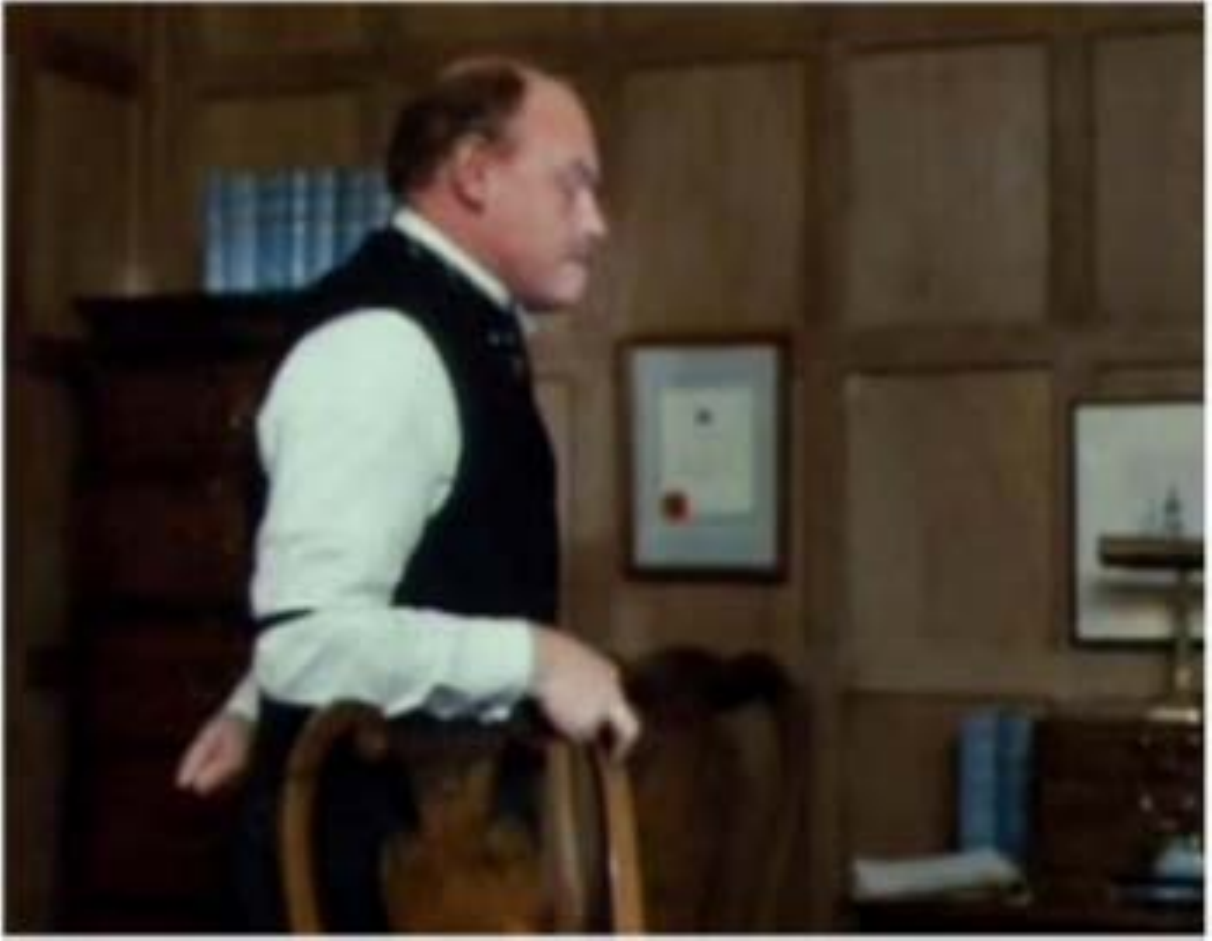}
\includegraphics[width=.16\textwidth]{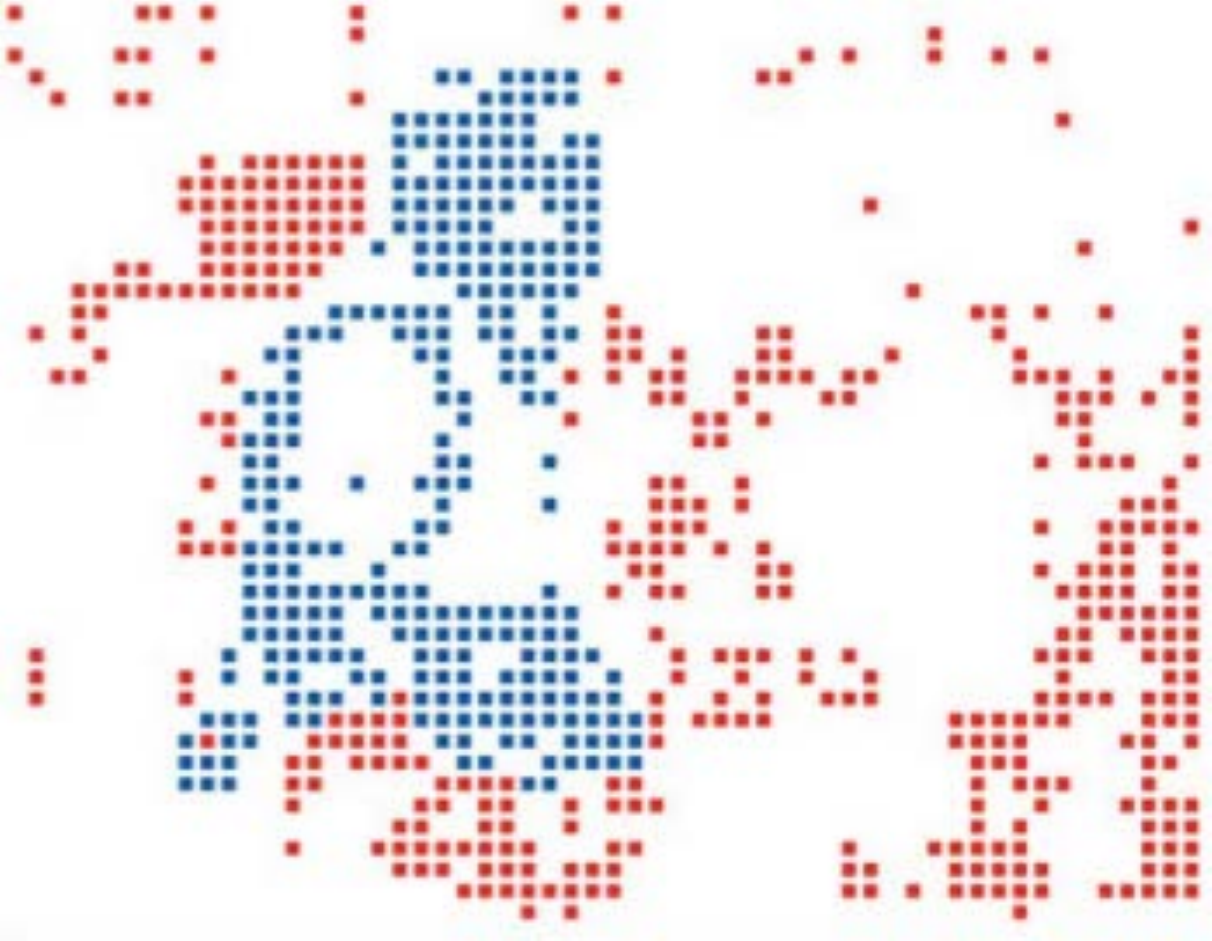}
\includegraphics[width=.16\textwidth]{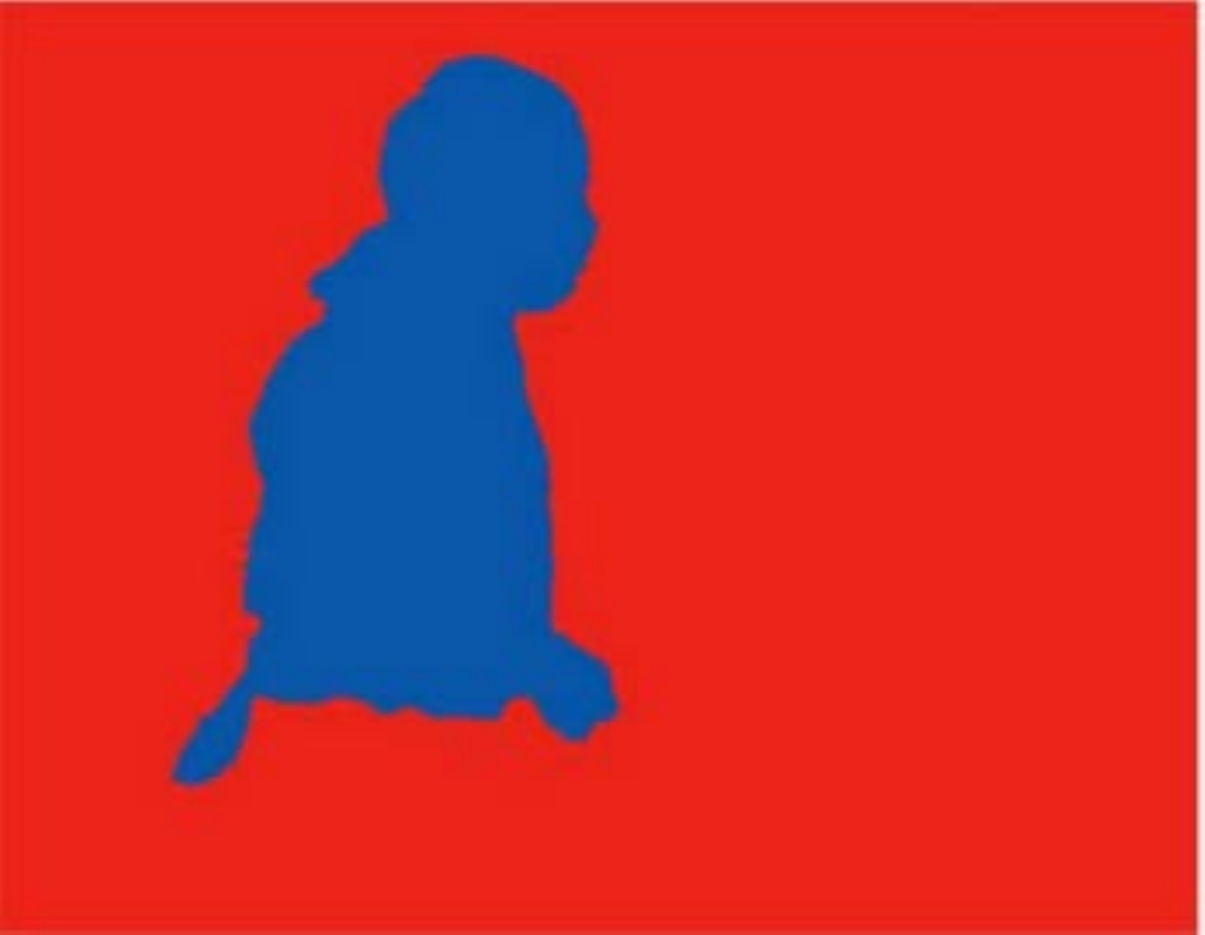}
\includegraphics[width=.16\textwidth]{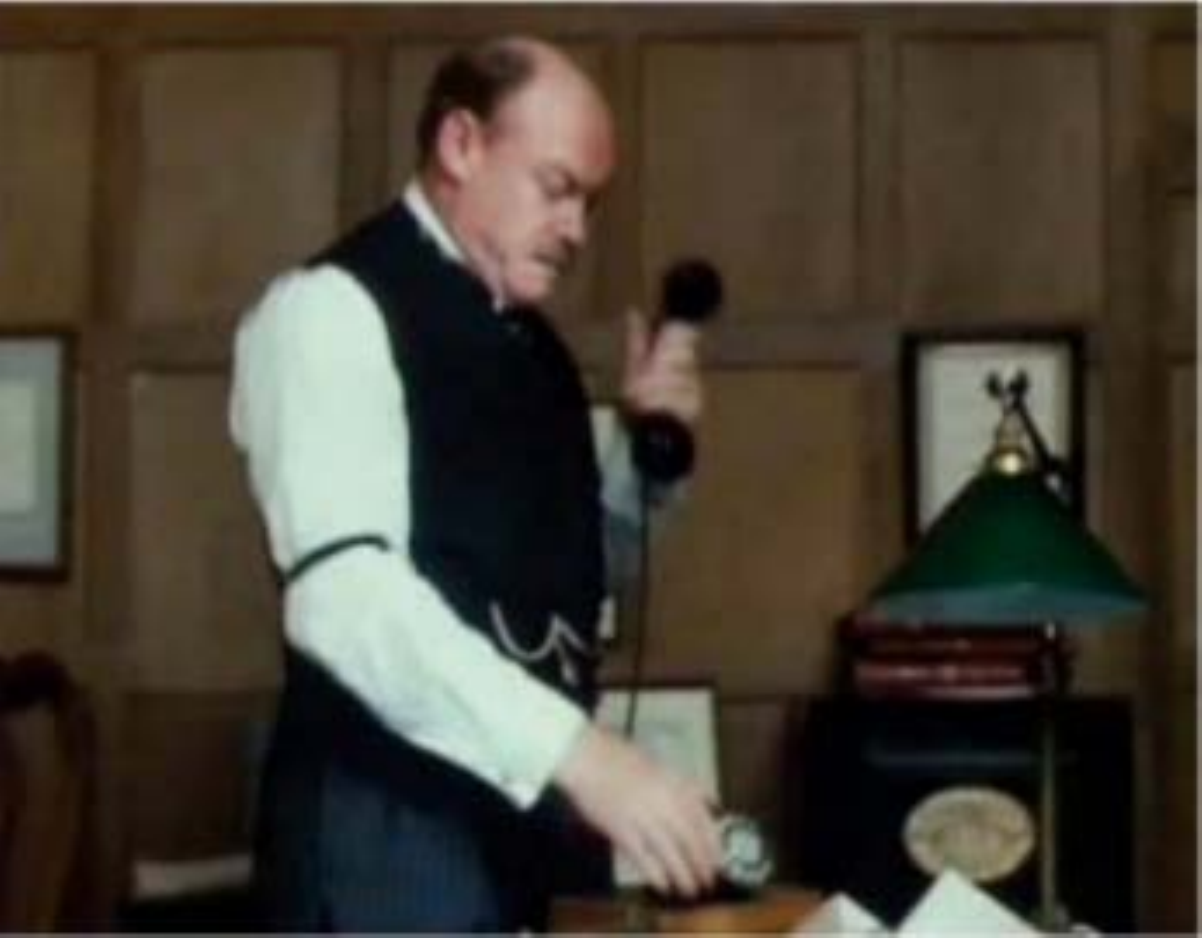}
\includegraphics[width=.16\textwidth]{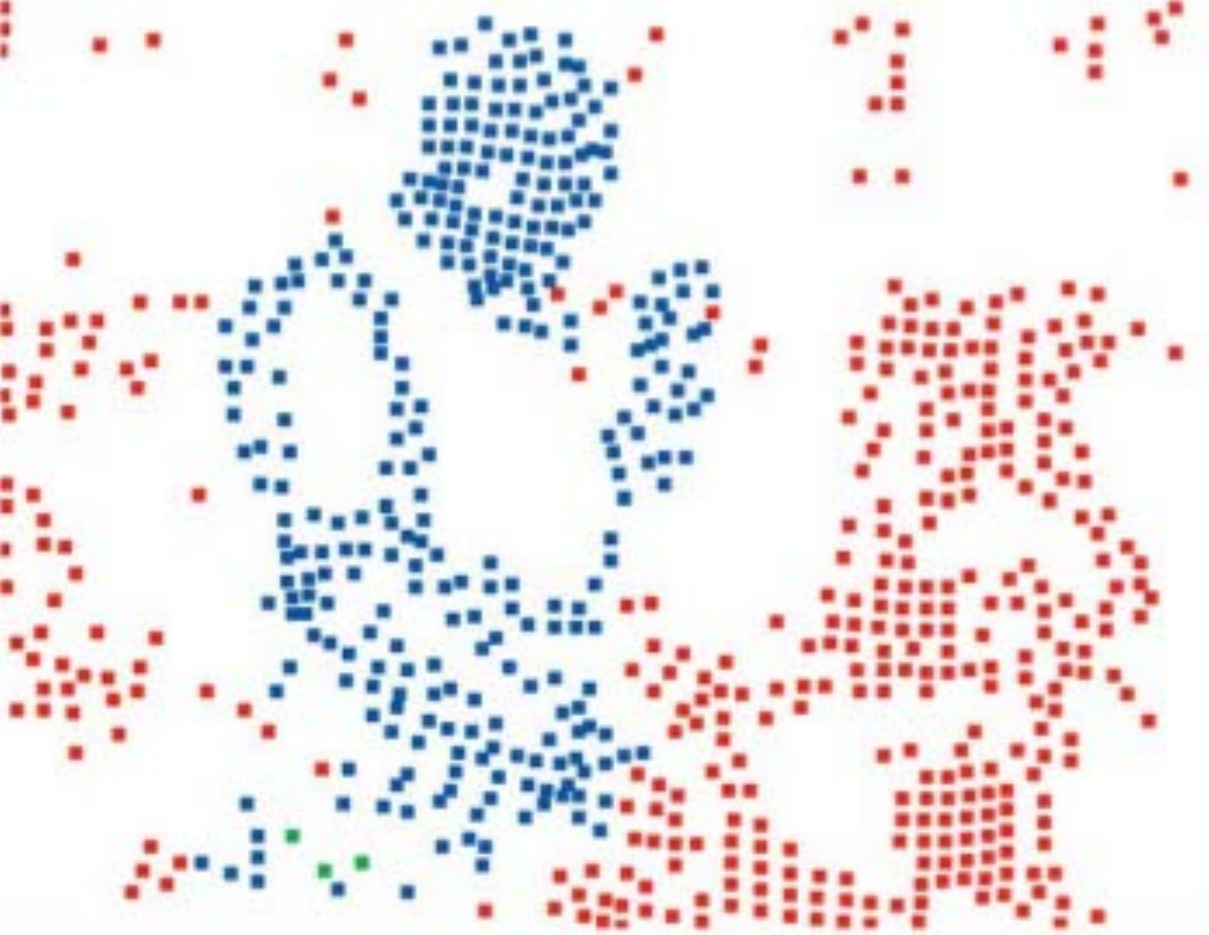}
\includegraphics[width=.16\textwidth]{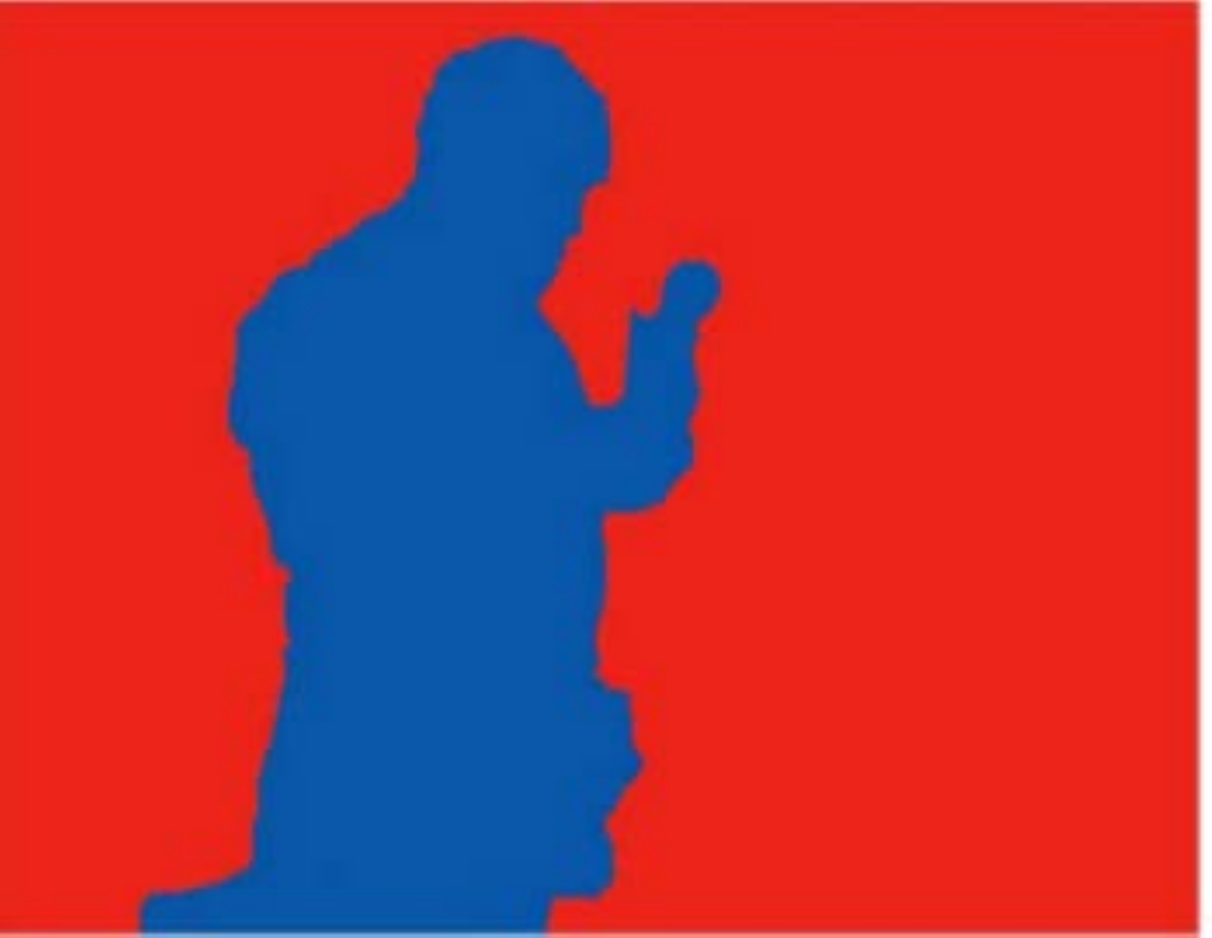}
\caption{Illustration of video object segmentation based on point trajectory \cite{Peter2014Segmentation}. From left to right: frame 1 and 4: a shot of video, image 2 and 5: clustering of point, image 3 and 6: the results of segmentation. }
\label{fig:point}
\end{figure}

The problem of video object segmentation can be solved by analyzing motion information over longer period. Motion is a strong perceptual cue for segmenting a video into separate objects~\cite{Shi1998}. Works of~\cite{Brox2010,6044588,ochs2012higher,fragkiadaki2012video,guizilini2013online,chen2015video,held2016probabilistic} approaches use long term motion information with point trajectories to take advantage of motion information available in multiple frames in recent years. Typically, these methods first generate point trajectories and then cluster the trajectories by using their affinity matrix. Finally, the clustering trajectory is used as the prior information to obtain the video object segmentation result. We divide point trajectory based VOS methods into two subcategories based on the motion estimation used, namely, optical flow and feature tracking methods. Optical flow estimates a dense motion field from one frame to the next, while feature tracking follows a sparse set of salient image points over many frames.

\paragraph{Optical flow based methods}
Many video object segmentation methods heavily rely on the dense optical flow estimation and motion tracking. 
Optical flow is a dense field displacement vector used to determine the pixels of each region. And it is usually used to capture spatio-temporal motion information of the objects of a video~\cite{horn1981determining}. There are two basic assumptions: 
\begin{itemize}
\item The brightness is constant. That is, when the same object moves between different frames, its brightness does not change. This is the assumption of the basic optical flow method for obtaining the basic equations of the optical flow method.
\item Small movement. That is, the change of time does not cause a drastic change in the object position, and the displacement between adjacent frames is relatively small.
\end{itemize}
Classic point tracking method use Kanade-Lucas-Tomasi~(KLT) \cite{lucas1981iterative} to generate sparse point trajectories. 
Brox and Malik~\cite{Brox2010} first perform point tracking to build sparse long-term trajectories and divide them into multiple clusters. Compared to the two-frame motion field, they argue that analyzing long-term point trajectories can better obtain temporally consistent clustering on many frames. In order to calculate such a point trajectory, this work runs a tracker developed in~\cite{sundaram2010dense} based on the large displacement optical flow~\cite{brox2011large}. As shown in Fig.~\ref{fig:point}, Ochs \emph{et al.}~\cite{Peter2014Segmentation} propose to use a semi-dense point tracker based on optical flow~\cite{Brox2010}, which can generate reliable trajectories of hundreds of frames with only a small drift and maintain a wide coverage of the video lens. Chen \emph{et al.}~\cite{chen2015video} employ both global and local information of point trajectories to cluster trajectories into groups.

\paragraph{Feature tracking based methods}
The above-mentioned methods first estimated the optical flow and then processed the discontinuities. In general, optical flow measurement is difficult in areas with very small textures or movements. In order to solve this problem, these methods propose to implement smoothing constraints in the flow field for interpolation. However, one must first know the segmentation to avoid the requirement of smooth motion discontinuity. Shi and Malik~\cite{Shi1998} define a motion feature vector at each pixel called motion profile, and adopt the normalized cuts~\cite{shi2000normalized} to divide a frame into motion segments. In~\cite{6044588}, spatio-temporal video segmentation algorithm is proposed to incorporate long-range motion cues from the past and future frames in the form of clusters of point tracks with coherent motion. Later, based on~\cite{Brox2010}, Ochs and Brox~\cite{ochs2012higher} introduce a variational method to obtain dense segmentations from such sparse trajectory clusters. Their method employs the spectral clustering to group trajectories based on a higher-order motion model. Fragkiadaki \emph{et al.}~\cite{fragkiadaki2012video} propose to detect discontinuities of embedding density between spatially neighboring trajectories. In order to segment the non-rigid object, they combine motion grouping cues to produce context-aware saliency maps. Moreover, a probabilistic 3D segmentation method~\cite{held2016probabilistic} is proposed to combine spatial, temporal, and semantic information to make better-informed decisions.

\emph{Discussion.} For background subtraction based methods, they explore motion information in short term and do not perform well when objects keep static in some frames. In contrast, point trajectories usually use long range trajectory motion similarity for video object segmentation.  Objects captured by trajectory clusters have proven to have a long time frame. However, non-rigid object or large motion lead to frequent pixel occlusions or dis-occlusions. Thus, in this case, point trajectories may be too short to be used. Additionally, these methods lack the appearance of the object appearance, \emph{i.e.} with only low-level bottom-up information.

\subsubsection{Over-segmentation}
\label{sec:oversegment}

\begin{figure}[t]
\centering
\includegraphics[width=.5\textwidth]{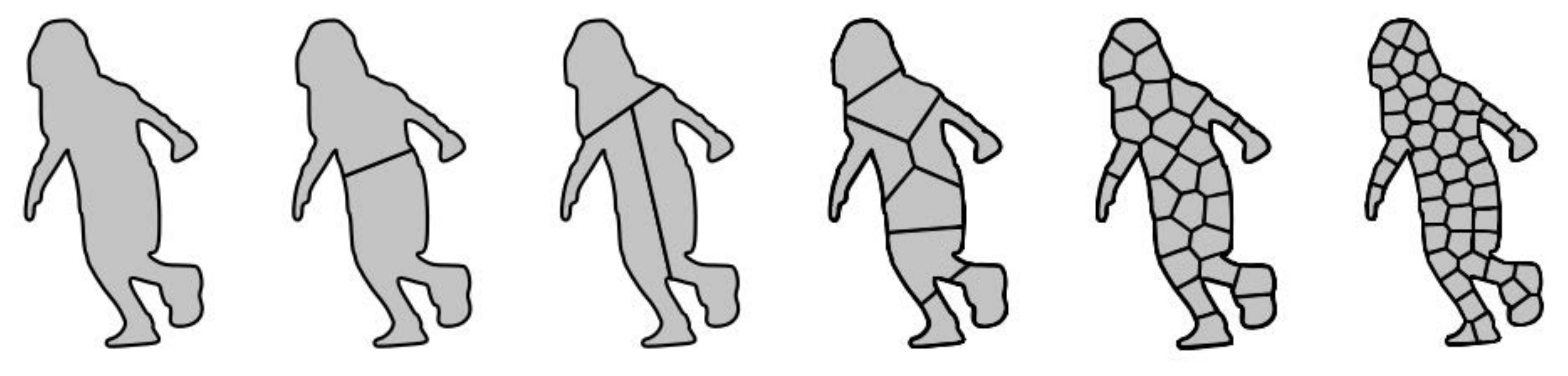}
\caption{Illustration of hierarchy of over-segmentations~\cite{chang2013video}. From over-segmentation of a human contour  (left) to superpixel (right).}
\label{fig:oversegment}
\end{figure}

Several over-segmentation approaches group pixels based on color, brightness, optical flow, or texture similarity and produce spatio-temporal segmentation maps~\cite{stein2007learning,yang2017video}. These methods generate an over-segmentation of the video into space-time regions.  
Fig.~\ref{fig:oversegment} shows the valid over-segmentations for a given human image. This example illustrates the difference between oversegmentations and superpixels. Although the terms ``over-segmentation'' and ``super-pixel segmentation'' are often used interchangeably, there are some differences between them. A superpixel segmentation is an over-segmentation that preserves most of the structure necessary for segmentation.

However, the number of over-segmentation makes optimization over sophisticated models intractable. Most current methods for unsupervised video object segmentation are graph-based~\cite{grundmann2010efficient,levinshtein2012optimal,jang2016primary}. Graph-based approaches to pixel, superpixel, or supervoxel generation treat each pixel as a node in a graph, where the vertices $\mathbf{D}=\{p,q,...\}$ of a graph $\mathbf{G}$ are partitioned into $N$ disjoint subgraphs, $\mathbf{A}_i, \mathbf{D}=\cup_{i=1}^N \mathbf{A}_i, \mathbf{A}_i\cap \mathbf{A}_j = \emptyset, i \neq j$, by pruning the weighted edges of the graph. The task is to assign a label $l\in \mathbf{L}$ to each $p\in \mathbf{D}$, the vertices are created by minimizing a cost function defined over the graph. Given a set of vertices $\mathbf{D}$ and a finite set of labels $\mathcal{L}$, an energy function is:
\begin{align}
E(f) = \sum_{p\in \mathbf{D}}U_p(f_p) + \lambda \sum_{\{p,q\}\in \mathcal{N}} w_{pg}\cdot V_{pg}(f_p,f_q),
\end{align}
where the unary term $U_p(f_p)$ express how likely is a label $l$ for pixel $p$, the pairwise term $V_{pg}(f_p,f_q)$ represent how likely labels $l_1$ and $l_2$ are for neighboring pixels $p$ and $q$, and $\mathcal{N}$ is a collection of neighboring pixel pairs. The coefficients $w_{pg}$ are the weights, and $\lambda$ is the parameter.

\begin{center}
\begin{table}
\caption{Summary of some major unsupervised VOS methods. \#: number of objects, S: single, M: multiple.}
\label{tab:unsupervised1} 
\scalebox{0.91} 
{
\begin{tabular}
{p{2.2cm}|c|p{2.2cm}|c|p{7cm}}
\hline
References & \# & Features & Optical flow & Methods \\
\hline  \hline
\multicolumn{5}{c}{Background subtraction methods} \\
\hline 
Han~\cite{han2012density} & M & RGB color, gradient, Haar-like & $\times$ & Stationary backgrounds: generative (KDA) and discriminative(SVM) techniques \\
\hline
Stauffer~\cite{stauffer2000learning} & M & RGB color & $\times$ & Stationary backgrounds: model each pixel as MoG and use on-line approximation to update \\
\hline
Criminisi~\cite{criminisi2006bilayer} & M & YUV color & $\times$ & 2D motion: probabilistic fusion of motion, color and contrast cues using CRF \\
\hline
Ren~\cite{ren2003statistical} & M & RGB color & $\times$ & 2D motion: estimate motion compensation using Gaussians model \\
\hline
Torr~\cite{torr1998concerning} & M & Corner features & $\times$ & 3D motion: a maximum likelihood and EM approach to clustering \\
\hline
Irani~\cite{irani1998unified} & M & RGB color & $\times$ & 3D motion: unified approach to handling moving-object detection in both 2D and 3D
scenes \\
\hline
\multicolumn{5}{c}{Point trajectory methods} \\
\hline 
Ochs~\cite{Peter2014Segmentation} & M & RGB color & $\surd$ & Optical flow: build spare long-term trajectories using optical flow\\
\hline
Fragkiadaki~\cite{fragkiadaki2012video} & M & RGB color & $\surd$ & Optical flow: combine motion grouping cues with context-aware saliency maps \\
\hline
Held~\cite{held2016probabilistic} & M & Centroid, shape & $\times$ & Feature tracking: probabilistic framework to combine spatial, temporal, and semantic information \\
\hline 
Lezama~\cite{6044588} & M & RGB color & $\times$ & Feature tracking: spatio-temporal graph-based video segmentation with long-rang motion cues \\
\hline
\multicolumn{5}{c}{Over-segmentation methods} \\
\hline
Giordano~\cite{giordano2015superpixel} & M & RGB color & $\times$ & Superpixel: generate coarse segmentation with superpixel and refine mask with energy functions \\
\hline
Chang~\cite{chang2013video} & M & LAB color, image axes & $\surd$ & Supervoxel: object parts in different frames, model flow between frames with a bilateral Gaussian process \\
\hline
Grundmann~\cite{grundmann2010efficient} & M & \small{$\chi^2$ distance of the normalized color histograms} & $\times$ & Supervoxel: graph-based method, iteratively process over multiple levels \\
\hline
Stein~\cite{stein2007learning} & M & Brightness and color gradient & $\times$ & Boundaries: graphical model, learn local classifier and global inference
\\
\hline
\multicolumn{5}{c}{Convolutional neural network methods} \\
\hline
Tokmakov~\cite{tokmakov2017learning} & S & Deep features & $\surd$ & Build instance embedding network and link objects in video \\
\hline
Jain~\cite{dutt2017fusionseg} & S & Deep features & $\surd$ & Design a two-stream fully CNN to combine appearance and motion information \\
\hline
Tokmakov~\cite{tokmakov2017learningmemory} & S & Deep features & $\surd$ & Trains a two-stream network RGB and optical flow, then feed into ConvGRU \\
\hline
Vijayanarasimha \cite{vijayanarasimhan2017sfm} & M & Deep features & $\surd$ & Geometry-aware CNN to predict depth, segmentation, camera and rigid object motions \\
\hline
Tokmakov~\cite{tokmakov2017learning} & S & Deep features & $\surd$ & Encoder-decoder architecture, coarse representation of the optical flow, then refines it iteratively to produce motion labels\\
\hline
Song~\cite{song2018pyramid} & S & Deep features & $\times$ & Pyramid dilated bidirectional ConvLSTM architecture, and CRF-based post-process \\
\hline
\multicolumn{5}{r}{\textit{Continued on next page}}
\end{tabular}
}
\end{table}
\end{center}

\setcounter{table}{0}
\begin{center}
\begin{table}[t]
\caption{\textit{Continued on next page.} ACC: Area, centroid, average color. HOOF: Histogram of Oriented Optical Flow. NG: Objectness via normalized gradients. HOG: Histogram of oriented gradients.}
\label{tab:unsupervised2} 
\scalebox{0.91} 
{
\begin{tabular}
{p{2.2cm}|c|p{2.2cm}|c|p{7cm}}
\hline
References & \# & Features & Optical flow & Methods \\
\hline  \hline
\multicolumn{5}{c}{``Object-like'' segments methods} \\
\hline
Li~\cite{Li2014Video} & M & Bag-of-words on color SIFT & $\surd$ & Segment: generate a pool of segment proposals, and online update the model \\
\hline
Lee~\cite{lee2011key} & S & $\chi^2$ distance between unnormalized color histograms &  $\surd$ & discover key-segments, space-time MRF for foreground object segmentation \\
\hline
Koh~\cite{Koh_2018_ECCV} & M & Deep features & $\surd$ & Saliency: extract object instances using sequential clique optimization \\
\hline
Wang~\cite{wang2015saliency} & S & Edges, motion & $\surd$ & Saliency: generate frame-wise spatio-temporal saliency maps using geodesic distance \\
\hline
Faktor~\cite{faktor2014video} & S & RGB color, local structure & $\surd$ & Saliency: combine motion saliency and visual similarity across large time-laps
\\
\hline
Wang~\cite{ChaohuiWang2009} & M & RGB color & $\times$ & Object proposals: MRF for segmentation, depth ordering and tracking with occlusion handling \\
\hline
Perazzi~\cite{Perazzi2015} & S & ACC, HOOF, NG, HOG & $\surd$ & Object proposals: geodesic superpixel edge Fully connected CRF using multiple object proposals \\
\hline
Tsai~\cite{tsai2016video} & S & Color GMM, CNN feature & $\surd$ & Object proposals: CRF for joint optimization of segmentation and optical flow \\
\hline
Koh~\cite{koh2017primary} & S & Bag-of-visual-words with LAB color & $\surd$ & Object proposals: ultrametric contour maps (UCMs) in each frame, refine primary object regions \\
\hline
\end{tabular}
}
\end{table}
\end{center}

\paragraph{Superpixel representation}
Shi and Malik~\cite{shi2000normalized} propose the graph-based normalized cut to overcome the oversegmentation problem.
The term superpixel is coined by Ren and Malik~\cite{ren2003learning} in their work on learning a binary classifier that can segment natural images. They use the normalized cut algorithm~\cite{shi2000normalized} for extracting the superpixels, with contour and texture cues incorporated. Levinshtein \emph{et al.}~\cite{levinshtein2012optimal} introduce the concept of spatio-temporal closure, and automatically recovers coherent components in images and videos, corresponding to objects and object parts. Chang~\emph{et al.}~\cite{chang2013video} develop a graphical model for temporally consistent superpixels in video sequences, and propose a set of novel metrics to quantify performance of a temporal superpixel representation: object segmentation consistency, 2D boundary accuracy, intra-frame spatial locality, inter-frame temporal extent, and inter-frame label consistency. Giordano~\emph{et al.}~\cite{giordano2015superpixel} generate a coarse foreground segmentation to provide predictions about motion regions by analyzing the superpixel segmentation changes in consecutive frames, and refine the initial segmentation by optimizing an energy function. In~\cite{yang2016fast}, appearance modeling technique with superpixel for automatic primary video object segmentation in the Markov random field (MRF) framework is proposed. Jang~\emph{et al.}~\cite{jang2016primary} introduce three foreground and background probability distributions: Markov, spatio-temporal, and antagonistic to minimize a hybrid of these energies to separate a primary object from its background. Furthermore, they refine the superpixel-level segmentation results. Yang~\emph{et al.}~\cite{yang2017video} introduce a multiple granularity analysis framework to handle a spatio-temporal superpixel labeling problem.

\paragraph{Supervoxel representation}
Several supervoxel-the video analog to a superpixel-methods over-segment a video into spatio-temporal regions of uniform motion and appearance~\cite{grundmann2010efficient,oneata2014spatio}. Grundmann~\emph{et al.}~\cite{grundmann2010efficient} over-segment a volumetric video graph into space-time regions grouped by appearance, and propose a hierarchical graph-based algorithm for spatio-temporal segmentation of long video sequences. In~\cite{oneata2014spatio}, Oneata~\emph{et al.} build a 3D space-time voxel graph to produce spatial, temporal, and spatio-temporal proposals by a randomized supervoxel merging process, and the algorithm is based on an extremely fast superpixel algorithm: simple linear iterative clustering (SLIC)~\cite{achanta2012slic}.

In addition to superpixel representation, there are some other video object segmentation methods based on over-segmentation, such as boundaries~\cite{wang1998unsupervised,stein2007learning}, patches~~\cite{huang2009video,schiegg2014graphical}. Wang~\cite{wang1998unsupervised} proposes a unsupervised video segmentation method with spatial segmentation, marker extraction, and modified watershed transformation. The algorithm partitions the first frame into homogeneous regions based on intensity, motion estimation to estimate motion parameters for each region. Stein~\emph{et al.}~\cite{stein2007learning} present a framework for introducing motion as a cue in detection and grouping of object or occlusion boundaries. A hypergraph cut method~\cite{huang2009video} is proposed to over-segment each frame in the sequence, and take the over-segmented image patches as the vertices in the graph. Schiegg~\emph{et al.}~\cite{schiegg2014graphical} build an undirected graphical model that couples decisions over all of space and all of time, and joint segment and track a time-series of oversegmented images/volumes for multiple dividing cells.

\emph{Discussion.} In general, the over-segmentation approaches occupy the space between single pixel matching and standard segmentation approaches. The algorithm reduce the computational complexity, since disparities only need to be estimated per-segment rather than per-pixel. However, in more complex videos, the over-segmentation requires additional knowledge, and are sensitive to boundary strength fluctuations from frame to frame. 

\subsubsection{``Object-like'' segments}
\label{sec:object}

Several recent approaches aim to upgrade the low-level grouping of pixels (such as pixel, superpixel, and supervoxel) to \emph{object-like} segments~\cite{fukuchi2009saliency,sundberg2011occlusion,Papazoglou2014Fast}. Although the details are different, the main idea is to generate a foreground object hypothesis for each frame of the image using the learning model of the "object-like" regions (such as salient objects, and object proposals from background).

\paragraph{Salient objects}
In~\cite{rahtu2010segmenting,wang2015saliency,hu2018unsupervised}, these works introduce saliency information as prior knowledge to discover visually important objects in a video. A spatio-temporal video modeling and segmentation method is proposed to partition the video sequence into homogeneous segments with selecting salient frames~\cite{song2007selecting}. Fukuchi \emph{et al.}~\cite{huang2009video} propose a automatic video object segmentation method based on visual saliency with the maximum a posteriori (MAP) estimation of the MRF with graph cuts. Rahtu~\emph{et al.} \cite{rahtu2010segmenting} present a salient object segmentation method based on combining a saliency measure with a Conditional Random Field (CRF) model using local feature contrast in illumination, color, and motion information. Papazoglou and Ferrari~\cite{Papazoglou2014Fast} compute a motion saliency map using optical flow boundaries, and handle fast moving backgrounds and objects exhibiting a wide range of appearance, motions and deformations. Faktor and Irani~\cite{faktor2014video} perform saliency votes at each pixel, and iteratively correct those votes by consensus voting of re-occurring regions across the video sequence to separate a segment track from the background. In~\cite{wang2015saliency}, Wang~\emph{et al.} produce saliency results via the geodesic distances to background regions in the subsequent frames, and build global appearance models for foreground and background based the saliency maps. Hu~\cite{hu2018unsupervised} propose a saliency estimation method and a neighborhood graph based on optical flow and edge cues for unsupervised video object segmentation. Koh~\emph{et al.}~\cite{Koh_2018_ECCV} generate object instances in each frame and develop the sequential clique optimization algorithm to consider both the saliency and similarity energies, then convert the tracks into video object segmentation results.

\paragraph{Object proposals}
Recently, video object segmentation methods generate object proposals in each frame, and then rank several object candidates to build object and background models~\cite{lee2011key,Li2014Video,xiao2016track,koh2017cdts,tsai2016video}. Typically, it contains three main categories: (i) figure-ground segmentations based object regions; (ii) optical flow based object proposals; (iii) bounding box based object proposals.

\begin{figure}[t]
\centering
\includegraphics[width=.3\textwidth]{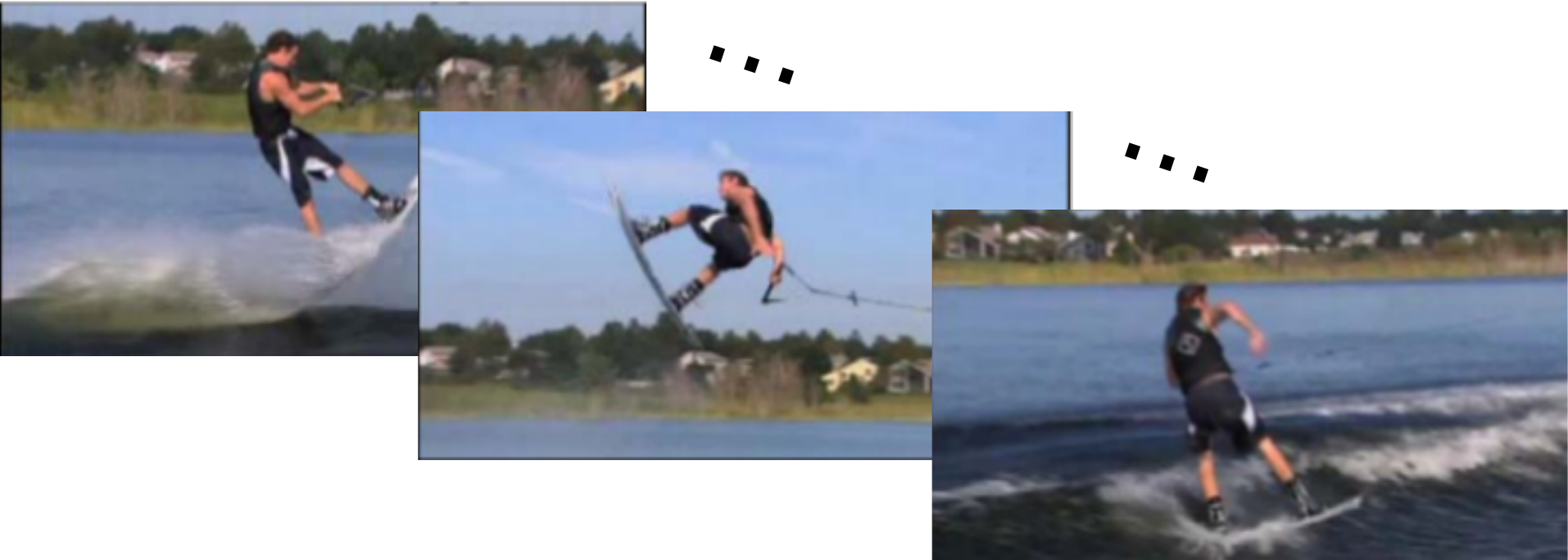} \hspace{5mm}
\includegraphics[width=.3\textwidth]{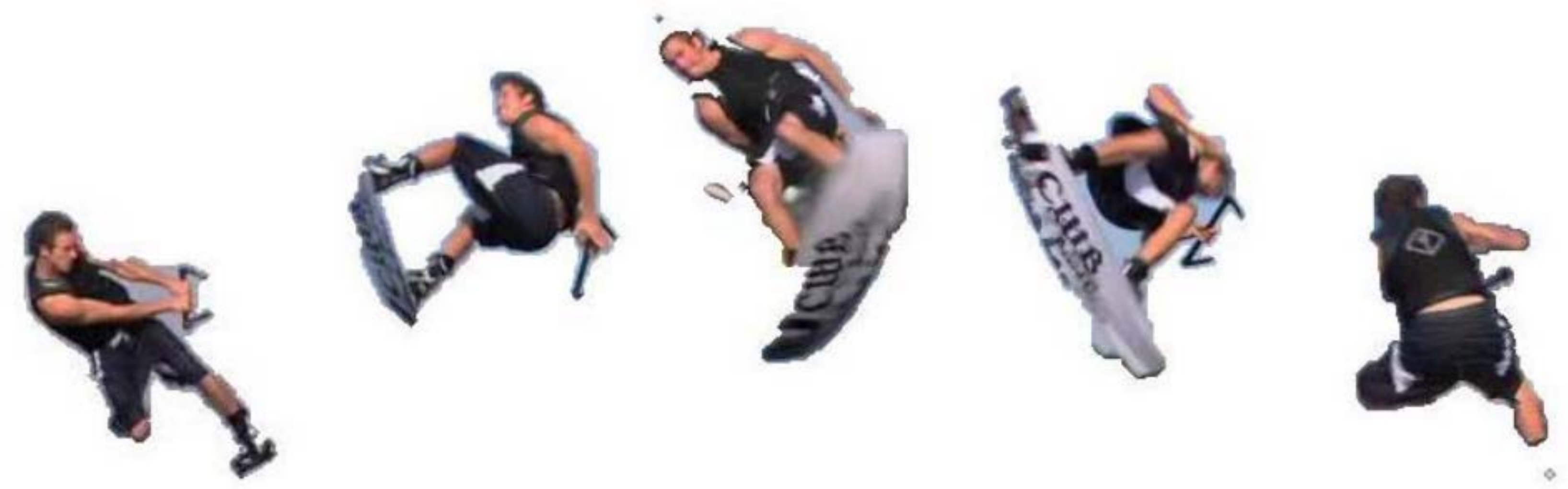}
\caption{Illustration of a set of key-segments to generate a foreground object segmentation of the video~\cite{lee2011key}. Left: video frames. Right: Foreground object segments.}
\label{fig:keysegment}
\end{figure}

For (i), the proposals are obtained by multiple static figure-ground segmentations similar to \cite{carreira2012cpmc,endres2010category}. Their works generate the generic foreground object using several image cues such as color, texture, and boundary. Brendel and Todorovic~\emph{et al.}~\cite{Brendel2009} segment a set of regions by using any available low-level segmenter in each frame, and cluster the similar regions across the video. In~\cite{lee2011key}, Lee~\emph{et al.} discover key-segments and group them to predict the foreground objects in a video. As shown in Fig.~\ref{fig:keysegment}, their works generate a diverse set of object proposals or key-segments in each frame using the static region-ranking method of \cite{endres2010category}. Later, Ma and Latercki produce a bag of object proposals in each frame using~\cite{endres2010category}, and build a video object segmentation algorithm using maximum weight cliques with mutex constraints. Banica~\emph{et al.}~\cite{banica2013video} generate multiple figure-ground segmentations based on boundary and optical flow cues, and construct multiple plausible partitions corresponding to the static and the moving objects. Moreover, Li~\emph{et al.}~\cite{Li2014Video} produce a pool of segment proposals using the figure-ground segmentation algorithm, and present a new composite statistical inference approach for refining the obtained segment tracks. To handle occlusion of the method~\cite{Li2014Video}, Wu~\emph{et al.}~\cite{wu2015robust} propose a video segment proposal approach start segments from any frame and track them through complete occlusions. Koh and Kim~\cite{koh2017primary} generate candidate regions using both color and motion edges, estimate initial primary object regions, and augment the initial regions with missing parts or reducing them. In addition, visual semantics~\cite{tsai2016semantic}, 
or context~\cite{wang2016primary} cues are used to generate object proposals and infer the candidates for subsequent frames.

For (ii), instead of modeling the foreground regions of statics object proposals, video object segmentation methods were presented by optical flow based object proposals~\cite{fragkiadaki2015learning,zhang2013video}. Lalos \emph{et al.}~\cite{lalos2010object} propose the object flow to estimate both the displacement and the direction of an object-of-interest. However, their work does not solve the problem of flow estimation and segmentation. Therefore, 
Fragkiadaki~\emph{et al.}~\cite{fragkiadaki2015learning} present a method to generate moving object proposals from multiple segmentations on optical flow boundaries, and extend the top ranked segments
into spatio-temporal tubes using random walkers. Zhang~\emph{et al.}~\cite{zhang2013video} present a optical flow gradient motion scoring function for selection of object proposals to discriminate between moving objects and the background. 

For (iii), several methods~\cite{koh2017cdts,xiao2016track} employ bounding box based spatio-temporal object proposals to segment video object recently. Xiao and Lee~\cite{xiao2016track} present a unsupervised algorithm to generate a set of spatio-temporal video object proposals boxes and pixel-wise segmentation. Koh and Kim~\cite{koh2017cdts} use object detector and tracker to generate multiple bounding box tracks for objects, transform each bounding box into a pixel-wise segment, and refine the segment tracks.

In addition, many researchers have exploited the Gestalt principle of ``common fate''~\cite{koffka2013principles} where similarly moving points are perceived as coherent entities, and grouping based on motion pointed out occlusion/disocclusion phenomena. In~\cite{sundberg2011occlusion}, Sundberg~\emph{et al.} exploit motion cues and distinguishes occlusion boundaries from internal boundaries based on optical flow to detect and segment foreground object. Taylor~\emph{et al.}~\cite{taylor2015causal} infer the long-term occlusion relations in video, and used within a convex optimization framework to segment the
image domain into regions. Furthermore, a video object segmentation method detect disocclusion in video of 3D scenes and to partition the disoccluded regions in objects.

\emph{Discussion.} Object-like regions (such as salient objects and object proposals) have been very popular as a preprocessing step for video object segmentation problems. Holistic object proposals can often extract over entire objects with optical flow, boundary, semantics, shape and other global appearance features, lead to better video object segmentation accuracy. However, these methods often generate many false positives, such as background proposals, which reduces segmentation performance. Moreover, these methods usually require heavy computational loads to generate object proposals and associate thousands of segments.

\subsubsection{Convolutional neural networks (CNN)} 
\label{sec:cnn}

Prior to the impressive of deep CNNs, some methods segment video object to rely on hand-crafted feature and do not leverage a learned video representation to build the appearance model and motion model. 

Recently, there have been attempts to build CNNs for video object segmentation.
The early primary video object segmentation method first generate salient objects using complementary convolutional neural network~\cite{li2017primary}, then propagate the video objects and superpixel-based neighborhood reversible flow in the video. Later, several video object segmentation methods employ deep convolutional neural networks in an end-to-end manner. In~\cite{tokmakov2017learning,tokmakov2017learningmemory,dutt2017fusionseg,vijayanarasimhan2017sfm}, these methods build a dual branch CNN to segment video object. MP-Net~\cite{tokmakov2017learning} takes the optical flow field of two consecutive frames of a video sequence as input and produces per-pixel motion labels. In order to solve the limitations of appearance features of object of MP-Net framework, Tokmakov~\emph{et al.}~\cite{tokmakov2017learningmemory} integrate one stream with appearance information and a visual memory module based on convolutional Gated Recurrent Units (GRU)~\cite{xingjian2015convolutional}. FSEG~\cite{dutt2017fusionseg} also proposes a two-stream network with appearance and optical flow motion to train with mined supplemental data. SfM-Net~\cite{vijayanarasimhan2017sfm} combines two streams motion and structure to learn object masks and motion models without mask annotations by differentiable rending. Li~\emph{et al.} \cite{li2018instance} transfer transferring the knowledge encapsulated in image-based instance embedding networks, and adapt the instance networks to video object segmentation. In addition, they propose a motion-based bilateral network, then a graph cut model is build to propagate the pixel-wise labels. In~\cite{goel2018unsupervised}, a deep reinforcement learning methods is proposed to automatically detect moving objects with the relevant information for action selection. Recently, Song~\emph{et al.}~\cite{song2018pyramid} present a video salient object detection method using pyramid dilated bidirectional ConvLSTM architecture, and apply it to the unsupervised VOS. Then, based on the CNN-convLSTM architecture, Wang~\emph{et al.}~\cite{Wang_2019_CVPR} propose a  visual attention-driven unsupervised VOS model. Additional, they collect unsupervised VOS human attention data from DAVIS~\cite{perazzi2016benchmark}, Youtube-Objects~\cite{prest2012learning}, and SegTrack v2~\cite{Li2014Video} dataset.

\vspace*{-2mm}

\subsubsection{Discussion}

Without any human annotation, unsupervised methods take the foreground object segmentation on an initial frame automatically.  They do not require user interaction to specify an object to segment. In other words, these methods exploit information of saliency, semantics, optical flow, or motion to generate primary objects, and then propagate it to the remainder of the frames. However, these unsupervised methods are not able to segment a specific object due to motion confusions between different instances and dynamic background. Furthermore, the problem with these unsupervised methods is that they are computationally expensive due to many unrelated interference object-like proposals. A qualitative comparison of some major unsupervised VOS methods are listed in Tab.~\ref{tab:unsupervised1}.

\vspace*{-2mm}

\subsection{Semi-supervised video object segmentation}
\label{sec:semi}

Semi-supervised video object segmentation methods are given with an initial object mask in the first frame or key frames. Then, these methods segment the object in the remaining frames. Typically, it can be investigated in the following main two categories: spatio-temporal graph and CNN based semi-supervised VOS.



\subsubsection{Spatio-temporal graphs}
\label{sec:handcrafted}

In recent years, early methods often solve some spatio-temporal graph with hand-crafted feature representation including appearance, boundary, and optical flows, and propagate the foreground region in the entire video. These methods typically rely on two important cues: object representation of graph structure and spatio-temporal connections.

\paragraph{Object representation of graph structure.}
Typically, the task is formulated as a spatio-temporal label propagation problem, these methods tackle the problem by building up graph structures over the object representation of (i) pixels, (ii) superpixels, or (iii) object patches to infer the labels for subsequent frames.

For (i), the pixels can maintain very fine boundaries, and are incorporated into the graph structure for video object segmentation. Tsai~\emph{et al.}~\cite{Tsai2012} perform MRF optimization using pixel appearance similarity and motion coherence to separate a foreground object from the background. In~\cite{marki2016bilateral}, given some user input as a set of known foreground and background pixels, M{\"a}rki~\emph{et al.} design a regularly sampled spatio-temporal bilateral grid, and minimize implicitly approximates long-range, spatio-temporal connections between pixels. Several methods build the dense~\cite{wang2017super} and sparse trajectories~\cite{101007978} to segment the moving objects in video by using a probabilistic model.

\begin{table}[t]
\caption{Summary of spatio-temporal graphs of semi-supervised VOS methods. Object rep. denotes object representation.}
\label{tab:semi-graphs}
\centering
\scalebox{0.91}{
\begin{tabular}{p{2.2cm}|c|p{3.5cm}|p{4.3cm}|c}
\hline
References & Object rep. & Connections & Appearance features & Optical flow \\
\hline
\hline
Wang \cite{wang2017super} & Pixel & Dense point clustering & Spatial location, color, velocity & $\surd$ \\
\hline 
M{\"a}rki \cite{marki2016bilateral} & Pixel & Spatio-temporal lattices & Image axes, YUV color & $\times$ \\
\hline
Tsai \cite{tsai2016video} & Superpixel & CRF & RGB color, CNN & $\surd$ \\
\hline
Jiang \cite{jang2016semi} & Superpixel & MRF & LAB color & $\times$ \\
\hline
Perazzi \cite{Perazzi2015} & Patch & CRF & HOG & $\surd$ \\
\hline
Fan \cite{fan2015jumpcut} & Patch & Nearest neighbor fields & SURF, RGB color & $\surd$ \\
\hline
Jain \cite{Jain243} & Superpixel & MRF & Color histogram & $\times$ \\
\hline
Ramakanth \cite{avinash2014seamseg} & Patch & Nearest neighbor fields & Color histogram & $\times$ \\
\hline
Ellis \cite{101007978} & Pixel & Sparse point tracking & RGB color & $\times$ \\
\hline
Badrinarayanan \cite{badrinarayanan2013semi} & Patch & Mixture of tree & Semantic texton forests feature & $\times$ \\
\hline
Budvytis \cite{budvytis2012mot} & Superpixel & Mixture of tree & Semantic texton forests feature & $\times$ \\
\hline
Tsai \cite{Tsai2012} & Pixel & MRF  & RGB color & $\times$ \\
\hline 
Ren \cite{4270202} & Superpixel & CRF & Local brightness, color and texture & $\times$ \\
\hline 
Wang \cite{wang2004video} & Superpixel & Mean-shift & Image axes, RGB color & $\times$ \\
\hline 
Patras \cite{patras2003semi} & Patch & Maximization of joint probability & Image axes, RGB color & $\times$ \\
\hline 
\end{tabular}
}
\end{table}

For (ii), in order to suffer from high computational cost and noisy temporal links of pixel-based graphs, many methods extract superpixels at from the input frames, and construct the superpixel graph. Each node in the graph represents a label. An edge is added between any two adjacent neighbors.  Graph structures such as CRF~\cite{4270202,tsai2016video,WangUnderstanding}, MRF~\cite{Jain243}, or mixture of trees~\cite{budvytis2012mot} can be integrated into the framework to further improve the accuracy. For instance, Ren and Malik~\cite{4270202} generate a sets of superpixels in the images, and build a CRF to segment figure from background in each frame. Then figure/ground segmentation operates sequentially in each frame by utilizing both static image cues and temporal coherence cues. Unlike using probabilistic graphical model to segment images independently, other graphical models decompose each into spatial nodes, and seek the foreground-background label assignment that maximizes both appearance consistency and label smoothness in space and time. Jain and Grauman~\cite{Jain243} present a higher order spatio-temporal superpixel label consistency potential for video object segmentation. In~\cite{budvytis2012mot}, a mixture of trees model is presented to link superpixels from the first to the last frame, and obtain super-pixel labels and their confidence. In addition, some methods use mean shift~\cite{wang2004video} and random walker~\cite{jang2016semi} algorithm to separate the foreground from the background.

For (iii), to make video object segmentation more efficient, researchers embed per-frame object patches and employ different techniques to select to a set of temporally coherent segments by minimizing and energy function of spatio-temporal graph. For example, Ramakanth~\emph{et al.}~\cite{avinash2014seamseg} and Fan~\emph{et al.}~\cite{fan2015jumpcut} employ approximate nearest neighbor algorithm to compute a mapping between two framers or fields, then predict the labels. Tasi~\emph{et al.}~\cite{tsai2016video} utilize the CRF model to assign each pixel with a foreground or background label. Perazzi~\emph{et al.}~\cite{Perazzi2015} formulate as a minimization of a novel energy function defined over a fully connected CRF of object proposals, and use maximum a posteriori to inference the foreground-background segmentation. Following in the work in~\cite{budvytis2012mot}, Badrinarayanan~\emph{et al.}~\cite{badrinarayanan2013semi} propose a patch-based temporal tree model to link patches between frames. Patras~\emph{et al.}~\cite{patras2003semi} use watershed algorithm obtain the color-based segmentation in each frame, and employ an iterative local joint probability search algorithm to generates a sequence of label.

\paragraph{Spatio-temporal connections}
Another important cue is how to estimate temporal connections between nodes by using spatio-temporal lattices~\cite{marki2016bilateral}, nearest neighbor fields~\cite{avinash2014seamseg,fan2015jumpcut}, mixture of trees~\cite{budvytis2012mot,badrinarayanan2013semi}. Some methods even build up long-range connections using appearance-based methods~\cite{Perazzi2015,patras2003semi}. Besides the cue of the temporal connections, another important issue is selecting the solution of optimization algorithm. Some algorithms use the local greedy strategy to infer labels by considering only two or more adjacent frames at a time~\cite{fan2015jumpcut,avinash2014seamseg,marki2016bilateral,101007978}, while other algorithms try to find global optimal solutions considering all frames~\cite{Tsai2012,Perazzi2015}. The locally optimization strategies perform segmentation on-the-fly allowing for applications where data arrives sequentially, while globally optimal solutions solve the limitation of short range interactions. A brief summary of spatio-temporal graphs methods is shown in Tab.~\ref{tab:semi-graphs}.


\subsubsection{Convolutional neural networks}
\label{sec:cnn_semi}

With the success of convolutional neural networks on static image segmentation~\cite{long2015fully,lin2018exploring}, CNN based methods show overwhelming power when introduced to video object segmentation. According to the used techniques for temporal motion information, they can be grouped into two types: motion-based and detection-based.


\paragraph{Motion-based methods} In general, the motion-based methods utilize the temporal coherence of the object motion, and formulate the problem of mask propagation starting from the first frame or a given annotated frame to the subsequent frames. 

For (i), one class of methods are developed to train network to incorporate optical flow~\cite{cheng2017segflow,jampani2017video,li2018video,hu2018motion}. Optical flow is important in early stages of the video description. It is common to apply optical flow to VOS to maintain motion consistency. And optical flow represents how and where each and every pixel in the image is going to move in the future pipeline. These VOS methods typically use optical flow as a cue to track pixels over time to establish temporal coherence. For instance, SegFlow~\cite{cheng2017segflow}, MoNet~\cite{xiao2018monet}, PReMVOS~\cite{luiten2018premvos}, LucidTrack~\cite{LucidDataDreaming_CVPR17_workshops}, and VS-ReID~\cite{li2018video} methods consist of two branches: the color segmentation and the optical flow branch using the FlowNet~\cite{dosovitskiy2015flownet,Ilg_2017_CVPR}. To learn to exploit motion cues, these methods receive twice or triple inputs, including the target frame and two adjacent frames. Jampani~\emph{et al.}~\cite{jampani2017video} present a temporal bilateral network to propagate video frames in an adaptive manner by using optical flow as additional feature. With temporal dependencies established by optical flow, Bao~\emph{et al.}~\cite{8578724} propose a VOS method via inference in CNN-based spatio-temporal MRF. Hu~\emph{et al.}~\cite{hu2018motion} employ active contour on optical flow to segment moving object. 

To capture the temporal coherence, some methods employ a Recurrent Neural Network (RNN) for modeling mask propagation with optical flow~\cite{hu2017maskrnn,li2018video}. RNN has been adopted by many sequence-to-sequence learning problems because it is capable to learn long-term dependency
from sequential data. MaskRNN~\cite{hu2017maskrnn} build a RNN approach which fuses in each frame the output of a binary segmentation net and a localization net with optical flow. Li and Loy~\cite{li2018video} combine temporal propagation and re-identification functionalities into a single framework. 

\begin{figure}[t]
\centering
\includegraphics[width=.46\textwidth]{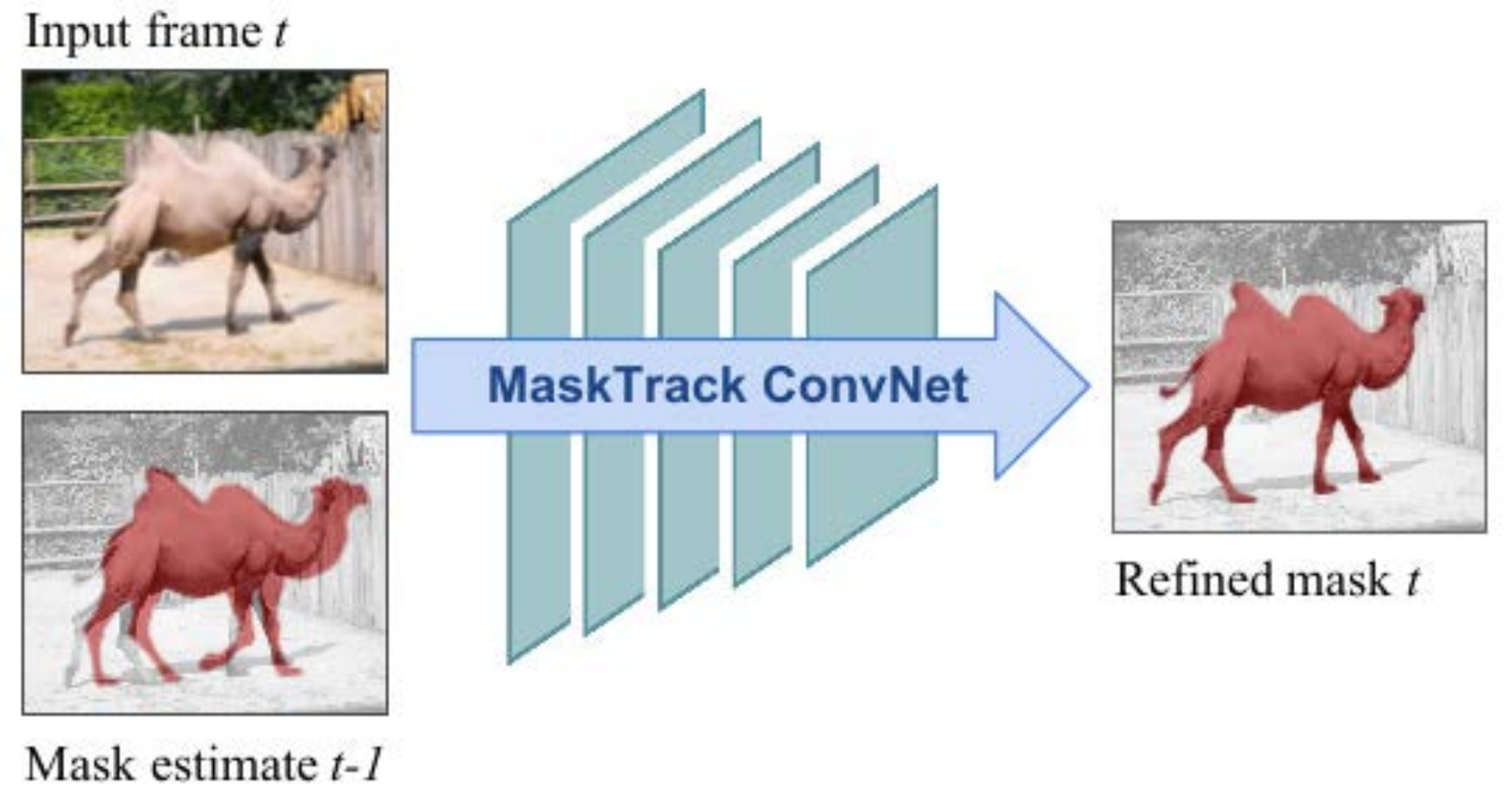} 
\caption{Illustration of mask prorogation to estimate the segmentation mask of the current frame from the previous frame as a guidance~\cite{perazzi2017learning}.}
\label{fig:masktrack}
\end{figure}

For (ii), another direction is to use CNNs to learn mask refinement of an object from current frame to the next one. For instance, as shown in Fig.~\ref{fig:masktrack}, MaskTrack~\cite{perazzi2017learning} method trains a refine the previous frame mask to create the current frame mask, and directly infer the results from optical flow. Compared to their approach~\cite{perazzi2017learning} that uses the exact foreground mask of the previous frame, Yang~\emph{et al.}~\cite{yang2018efficient} use a very coarse location prior with visual and spatial modulation. Oh~\emph{et al.}~\cite{wug2018fast} use both the reference frame with annotation and the current frame with previous mask estimation to a deep network. A reinforcement cutting-agent learning framework is to obtain the object box from the segmentation mask and propagates it to the next frame~\cite{han2018reinforcement}. Some methods leverage temporal information on the bounding boxes by tracking objects across frames~\cite{cheng2018fast,lee2018joint,newswanger2017one,sharir2017video}. Sharir~\emph{et al.}~\cite{sharir2017video} present a temporal tracking method to enforce coherent segmentation throughout the video. Cheng~\emph{et al.}~\cite{cheng2018fast} utilize a part-based tracking method on the bounding boxes, and construct a region-of-interest segmentation network to generate part masks. Recently, some methods~\cite{Xu_2018_ECCV,valipour2017recurrent} introduce a combination of CNN and RNN for video object segmentation. Xu~\emph{et al.}~\cite{Xu_2018_ECCV} generate the initial states for our convolutional Long Short-Term Memory (LSTM), and use a feed-forward neural network to encode both the first image frame and the segmentation mask. 

\paragraph{Detection-based methods} Without using temporal information, some methods learn a appearance model to perform a pixel-level detection and segmentation of the object at each frame. They rely on fine-tuning a deep network using the first frame annotation of a given test sequence~\cite{8100048,voigtlaender2017online}. Caelles~\emph{et al.}~\cite{8100048} introduce an offline and online training process by a fully convolutional neural network (FCN) on static image for one-shot video object segmentation (OSVOS), which fine-tunes a pretrained convolutional neural network on the first frame of the target video. Furthermore, they extend the model of the object with explicit semantic information, and dramatically improve the results~\cite{8362936}. Later, an online adaptive video object segmentation is proposed~\cite{voigtlaender2017online}, the network is fine-turned online to adapt to the changes in appearance. Cheng~\emph{et al.}~\cite{cheng2017learning} propose a method to propagate a coarse segmentation mask spatially based on the pairwise similarities in each frame.

Other approaches formulate video object segmentation as a pixel-wise matching problem to estimate an object of interest with subsequence images until the end of a sequence. Yoon~\emph{et al.}~\cite{shin2017pixel} propose a pixel-level matching network to distinguish the object area from the background on the basis of the pixel-level similarity between two object units. To solve computationally expensive problems, Chen~\emph{et al.}~\cite{chen2018blazingly} formulate a pixel-wise retrieval problem in an embedding space for video object segmentation, and VideoMatch approach~\cite{Hu_2018_ECCV} learns to match extracted features to a provided template without memorizing the appearance of the objects.


\paragraph{Discussion.} As indicated, CNN-based semi-supervised VOS methods can be roughly classified into: motion-based and detection-based ones. The classification of these two methods is based on temporal motion or non-motion. Temporal motion is an important feature cue in video object segmentation. As long as the appearance and position changes are smooth, the complex deformation and movement of the target can be handled. However, these methods are susceptible to temporal discontinuities such as occlusion and fast motion, and can suffer from drift once the propagation becomes unreliable.  
On the other hand, since such methods rarely rely on temporal consistency, they are robust to changes such as occlusion and rapid motion. However, since they need to estimate the appearance of the target, it is generally not possible to adapt to changes in appearance. It is difficult to separate the appearance of similar object instances.

A qualitative comparison of CNN-based semi-supervised VOS methods can be obtained based on motion-based or detection-based methods, requirement of optical flow, requirement of fine-tuning and computational speed, ability to handle post-processing, and requirement of data augmentation. In Tab.~\ref{tab:cnn_vos}, we provide the qualitative comparison of the methods discussed in this section.

\begin{figure}[t]
\centering
\includegraphics[width=.6\textwidth]{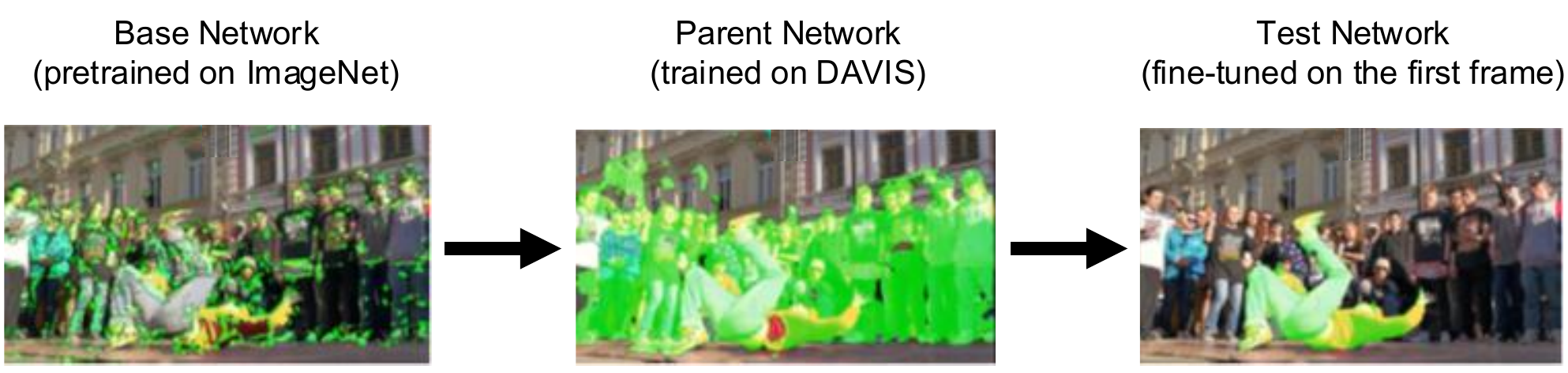} 
\caption{Illustration of the pipeline of one-shot video object segmentation~\cite{8100048}. The first step is to pertrain on large datasets (\emph{e.g.} ImageNet~\cite{deng2009imagenet} for image classification). The second step is to train parent network on the training set of DAVIS~\cite{perazzi2016benchmark}. Then, during test time, it fine-tunes on the first frame.}
\label{fig:cnn_semi_vos}
\end{figure}

\begin{itemize}
\item {\bf Fine-tuning.} Most of CNN-based semi-supervised VOS methods share a similar two-stage paradigm (as shown in Fig.~\ref{fig:cnn_semi_vos}): first, train a general-purpose CNN to segment the foreground object; second, this network use online fine-tuning using the first frame of the test video to memorize target object appearance, leading to a boost in the performance~\cite{8100048,voigtlaender2017online,perazzi2017learning}. It has been shown that fine-tuning on the first frame significantly improves accuracy. However, since at test time some methods only use the fine-tuned network, it is not able to adapt to large changes in appearance, which might for example be caused by drastic changes in viewpoint~\cite{8100048,8362936,hu2018motion,8578724,xiao2018monet,LucidDataDreaming_CVPR17_workshops}. And it becomes harder for the fine-tuned model to generalize to new object appearances. To overcome this limitation, some methods update the network online to changes in appearance using training examples~\cite{voigtlaender2017online,newswanger2017one}. 

\item {\bf Computational speed.} Despite the high accuracies achieved by these approaches, the fine-tuning process requires many iterations of optimization, the step on the video is computationally expensive, where it usually takes more than ten minutes to update a model and is not suitable for online vision applications. Recently, several methods~\cite{yang2018efficient,cheng2018fast,chen2018blazingly,Hu_2018_ECCV} work without the need of the computationally expensive fine-tuning in test time, and make them much faster than comparable methods. For instance, Chen~\emph{et al.}~\cite{chen2018blazingly} only perform a single forward through the embedding network and a nearest-neighbor search to process each frame in test time. Yang~\emph{et al.}~\cite{yang2018efficient} use a single forward pass to adapt the segmentation model to the appearance of a specific object. VideoMatch approach~\cite{Hu_2018_ECCV} build a soft matching layer, and does not require online fine-tuning.

\item {\bf Post-processing.} Besides the training of CNN-based segmentation, several methods leverage post-processing steps to achieve additional gains.
Post-processing is often employed to improve the contours, such as boundary snapping~\cite{8100048,8362936}, refine-aware filter~\cite{shin2017pixel,cheng2017learning},  and dense MRF or CRF~\cite{krahenbuhl2011efficient} in~\cite{8578724,perazzi2017learning}. For instance, OSVOS~\cite{8100048} perform boundary snapping to capture foreground masks to accurate contours. Yoon~\emph{et al.}~\cite{shin2017pixel} perform a weighted median filter on the resulting segmentation mask. Li~\emph{et al.}~\cite{li2018video} additionally consider post-processing steps to link the tracklets. In addition, 
some VOS frameworks~\cite{8578724,LucidDataDreaming_CVPR17_workshops,xiao2018monet} utilize MRF or CRF as a a post-processing step to improve the labeling results produced by a CNN. They attach the MRF or CRF inference to the CNN as a separate step, and utilize the representation capability of CNN and fine-grained probability modeling capability of MRF or CRF to improve performance. PReMVOS method~\cite{luiten2018premvos} present a refinement network that produces accurate pixel masks for each object mask proposal. 

\item {\bf Data augmentation.} In general, data augmentation is a widely strategy to improve generalization of neural networks. Khoreva \emph{et al.}~\cite{LucidDataDreaming_CVPR17_workshops} present a heavy data augmentation strategy for online learning. Other methods \cite{luiten2018premvos,jampani2017video,cheng2017segflow,sharir2017video} fine-tune the training network on a large set of augmented images generated from the first-frame ground truth.
\end{itemize}

\begin{table}[t]
\caption{Summary of convolutional neural network based semi-supervised video object segmentation methods. M/D: motion-based and detection-based methods. Post-pro.: post-processing. Data aug.: data augmentation.}
\label{tab:cnn_vos}
\centering
\scalebox{0.95}{
\begin{tabular}{l|c|c|c|c|c|c}
\hline 
References & M / D & Optical flow & Fine-tuning & Post-pro. & Speed & Data aug. \\
\hline \hline
Hu \cite{hu2018motion}  &  M & $\surd$ & $\surd$ & $\times$ &  & $\times$ \\
\hline
Bao \cite{8578724}  & M & $\surd$ & $\surd$ & $\surd$ &  & $\times$ \\
\hline
Xiao \cite{xiao2018monet} & M & $\surd$ & $\surd$ & $\surd$ &  & $\times$ \\
\hline
Khoreva \cite{LucidDataDreaming_CVPR17_workshops}  & M & $\surd$ & $\surd$ & $\surd$ &  & $\surd$ \\
\hline
Luiten \cite{luiten2018premvos}  & M & $\surd$ & $\surd$ & $\surd$ & fast & $\surd$ \\
\hline
Li \cite{li2018video}  & M & $\surd$ & $\times$ & $\surd$ &  & $\times$ \\
\hline
Yang \cite{yang2018efficient} & M & $\times$ & $\times$ & $\times$ & fast & $\times$ \\
\hline
Wug \cite{wug2018fast}  & M & $\times$ & $\surd$ & $\times$ & fast & $\times$ \\
\hline
Cheng \cite{cheng2018fast} & M & $\times$ & $\times$ & $\surd$ & fast & $\times$ \\
\hline
Han \cite{han2018reinforcement} & M & $\times$ & $\surd$ & $\times$ &  & $\times$ \\
\hline
Lee \cite{lee2018joint}  & M & $\times$ & $\surd$ & $\surd$ &  & $\times$ \\
\hline
Xu \cite{Xu_2018_ECCV}  & M & $\times$ & $\surd$ & $\times$ &  & $\times$ \\
\hline
Newswanger \cite{newswanger2017one}  & M & $\times$ & $\surd$ & $\surd$ &  & $\times$ \\
\hline
Sharir \cite{sharir2017video} & M & $\times$ & $\surd$ & $\times$ &  & $\surd$ \\
\hline 
Perazzi \cite{perazzi2017learning}  & M & $\times$ & $\surd$ & $\surd$ &   & $\surd$ \\
\hline 
Valipour \cite{valipour2017recurrent}  & M & $\times$ & $\times$ & $\times$ &  & $\times$ \\
\hline
Jampani \cite{jampani2017video}  & M & $\surd$ & $\times$ & $\times$ & fast & $\surd$ \\
\hline
Cheng \cite{cheng2017segflow}  & M & $\surd$ & $\surd$ & $\times$ &  & $\surd$ \\
\hline
Hu \cite{hu2017maskrnn}  & M & $\surd$ & $\surd$ & $\times$ &  & $\times$ \\
\hline
Maninis \cite{8362936}  & D & $\times$ & $\surd$ & $\surd$ &  & $\times$ \\
\hline 
Chen \cite{chen2018blazingly}  & D & $\times$ & $\times$ & $\times$ & fast & $\times$ \\
\hline
Hu \cite{Hu_2018_ECCV}  & D & $\times$ & $\times$ & $\times$ & fast & $\times$ \\
\hline
Caelles \cite{8100048}  & D & $\times$ & $\surd$ & $\surd$ &  & $\times$ \\
\hline
Voigtlaender \cite{voigtlaender2017online}  & D & $\times$ & $\surd$ & $\surd$ &  & $\times$ \\
\hline
Cheng \cite{cheng2017learning}  & D & $\times$ & $\surd$ & $\surd$ &  & $\times$ \\
\hline
Shin \cite{shin2017pixel}  & D & $\times$ & $\surd$ & $\surd$ &  & $\times$ \\
\hline
\end{tabular}
}
\end{table}

\subsection{Interactive video object segmentation}
\label{sec:interactive}

Interactive video object segmentation is a special form of supervised segmentation that relies on iterative user interaction to segment objects of interest. This is done by repeating the segmentation results of the correction system using additional strokes on the foreground or background. And these methods require the user to input either scribbles or clicks. In general, every segmentation algorithm needs to solve two problems, namely the criteria of good partitioning and the method of achieving effective partitioning~\cite{shi2000normalized}. Typically, interactive video object segmentation techniques can be divided into one of the following three main branches: graph partitioning models, active contours models, and convolutional neural network models.

\subsubsection{Graph partitioning models}
\label{sec:interactive_graph}

Most of image segmentation techniques of interactive video object segmentation methods are formulated as a graph partitioning problems, where the vertices $D$ of a graph $G$ are partitioned into disjoint $N$ subgraphs. Examples of existing interactive segmentation methods are graph-cuts, random walker, and geodesic based.
\paragraph{Graph-cuts based} Several works are based on the GrabCut algorithm~\cite{rother2004grabcut}, which iteratively alternates between estimating appearance models (typically Gaussian Mixture Models) and refining the segmentation using graph cuts~\cite{boykov2001interactive}. Wang \emph{et al.}\cite{wang2005interactive} were among the first authors to address interactive video segmentation tasks. To improve the performance, they used two-stage hierarchical mean-shift clustering as a preprocessing step to reduce the computation of the min-cut problem. In~\cite{li2005video}, Li~\emph{et al.} segment every tenth frame, and graph cut uses the global color model from keyframes, gradients and coherence as its primary clue to calculate the choice between frames. The user can also manually indicate the area in which the local color model is applied. Price~\emph{et al.}~\cite{price2009livecut} propose additional types of local classifiers, namely LIVEcut. The user iteratively corrects the propagated mask frame to frame and the algorithm learns from it. In~\cite{bai2009video}, Bai~\emph{et al.} build a set of local classifiers that each adaptively integrates multiple local image features. This method re-trains the classifier from the new mask by transforming the neighborhood regions according to the optical flow, and then retrains the user correction through the classifier. Later, the author construct the foreground and background appearance models adaptively in the same group~\cite{bai2010dynamic}, and use the probability optical flow to update the color space Gaussian of the individual pixel. In contrast to pixel, Reso~\emph{et al.}~\cite{reso2014interactive} and Dondera~\emph{et al.}~\cite{dondera2014interactive} adopt the graph-cut framework by using superpixels on every video frame. In addition, Chien~\emph{et al.}~\cite{Chien2013Video} and Pont-Tuset~\emph{et al.}~\cite{pont2015semi} use normalized cut~\cite{shi2000normalized} based multiscale combinatorial grouping (MCG) algorithm to segment and generate accurate region proposals, and use point clicks on the boundary of the objects to fit object proposals to them.

\paragraph{Random walker based}
In~\cite{shankar2015video}, Nagaraja~\emph{et al.} use a few strokes to segment videos by using optical flow and point trajectories. Their method integrate into a user interface where the user can draw scribbles in the first frame. When satisfied, the user presses a button to run the random walker.

\paragraph{Geodesic based}
Bai and Sapiro~\cite{bai2009geodesic} present a geodesics-based algorithm for interactive natural image and video by using region color to compute a geodesic distance to each pixel to form a selection. This method exploits weights in the geodesic computation that depend on the pixel value distributions. In~\cite{wang2017selective}, Wang~\emph{et al.} combine geodesic distance-based dynamic models with pyramid histogram-based confidence map to segment the image regions. Additionally, their method determines the frame of the operator's mark to improve segmentation performance.


\subsubsection{Active contours models}
\label{sec:interactive_active}

In the active contour framework, object segmentation use an edge detector to halt the evolution of the curve on the boundary of the desired object. Based on this framework, the TouchCut approach~\cite{wang2014touchcut} uses a single touch to segment the object using level-set techniques. They simplify the interaction to a single point in the first frame, and then propagates the results using optical flow.

\subsubsection{CNN models}
\label{sec:interactive_cnn}

Many recent works employ convolutional neural network models to accurately interactive segment the object in successive frames~\cite{maninis2018deep,chen2018blazingly,benard2017interactive,caelles20182018}. Benard~\emph{et al.}~\cite{benard2017interactive} and Caelles~\emph{et al.}~\cite{caelles20182018} propose the deep interactive image and video object segmentation method use OSVOS technique~\cite{8100048}. To improve localization, Benard~\emph{et al.} propose to refine the initial predictions with a fully connected CRF. Caelles~\emph{et al.}~\cite{caelles20182018} define a baseline method (\emph{i.e.} Scribble-OSVOS) to show the usefulness of the 2018 DAVIS challenge benchmark. Chen~\emph{et al.}~\cite{chen2018blazingly} formulate video object segmentation as a pixel-wise retrieval problem. And their method allow for a fast user interaction. iFCN~\cite{xu2016deep} guides a CNN from positive and negative points acquired from the ground-truth masks. In~\cite{maninis2018deep}, Maninis ~\emph{et al.} build on iFCN to improve the results by using four points of an object as input to obtain precise object segmentation for images and videos.

\begin{table}[t]
\caption{Summary of interactive video object segmentation methods. \#: number of objects, S: single, M: multiple.}
\label{tab:interacitve}
\centering
\scalebox{1}{
\begin{tabular}{p{2.5cm}|l|l|p{2.2cm}|l|l}
\hline 
References & \# & Methods & Way of labeling & Optical flow & Over-segmentation \\
\hline
\hline
Maninis~\cite{maninis2018deep} & M & CNN models & Clicks & $\times$ & Pixel \\ 
\hline
Caelles~\cite{caelles20182018} & M & CNN models & Scribbles & $\times$ & Pixel \\
\hline
Chen~\cite{chen2018blazingly} & M & CNN models & Clicks & $\times$ & Pixel \\
\hline
Benard~\cite{benard2017interactive} & S & CNN models & Clicks & $\times$ & Pixel \\
\hline 
Nagaraja~\cite{shankar2015video} & S & Random walker & Scribbles & $\surd$ & Pixel \\
\hline
Pont-Tuset\cite{pont2015semi} & S & Graph-cut & Clicks & $\surd$ & Superpixel \\
\hline
Chien~\cite{Chien2013Video} & S & Graph-cut & Clicks & $\times$ & Pixel \\
\hline
Wang~\cite{wang2014touchcut} & S & Active contours & Clicks & $\surd$  & Pixel \\
\hline
Donder~\cite{dondera2014interactive} & S & Graph-cut & Clicks & $\surd$  & Superpixel \\
\hline
Reso~\cite{reso2014interactive} & S & Graph-cut & Scribbles & $\surd$  & Superpixel \\
\hline
Bai~\cite{bai2010dynamic} & S & Graph-cut & Scribbles & $\surd$ & Pixel \\
\hline
Bai~\cite{bai2009geodesic} & S & Geodesic & Scribbles & $\surd$ & Pixel \\
\hline
Bai~\cite{bai2009video} & S & Graph-cut & Scribbles & $\surd$ & Pixel \\
\hline
Price~\cite{price2009livecut} & S & Graph-cut & Scribbles & $\times$ & Pixel \\
\hline
Li \cite{li2005video} & S & Graph-cut & Scribbles & $\times$ & Pixel \\
\hline
Wang~\cite{wang2005interactive} & S & Graph-cut & Scribbles & $\surd$ & Pixel \\
\hline
\end{tabular}
}
\end{table}

\subsubsection{Discussion.} Given scribbles or a few clicks by the user, the interactive video object segmentation helps the system produce a full spatio-temporal segmentation of the object of interest. Interactive segmentation methods have been proposed in order to reduce annotation time. However, on small touch screen devices, using a finger to provide precise clicks or drawing scribbles can be cumbersome and inconvenient for the user.
A qualitative comparison of interactive VOS methods can be made based on their ability to segment single or multiple objects, label an object with clicks or scribbles, and type of over-segmentation (\emph{i.e.} pixel or superpixel). A brief summary of the qualitative comparison is shown in Tab.~\ref{tab:interacitve}. Most of the conventional graph partitioning model based interactive VOS methods to seeded segmentation is the graph-cut algorithm. Recent methods use the ideas in the pipeline of deep architectures, CNN models are utilized to improve the interactive segmentation performance. In addition, several CNN models based methods can handle multiple-object interactive video object segmentation.

\subsection{Weakly supervised video object segmentation}
\label{sec:weakly}

Weakly supervised VOS can provide a large amount of video for this method, where all videos are known to contain the same foreground object or object class. Several weakly supervised learning-based approaches to generate semantic object proposals for training segment classifiers~\cite{hartmann2012weakly,tang2013discriminative} or performing label transfer~\cite{liu2014weakly}, and then produce the target object in videos. For instance, Hartmann~\emph{et al.}~\cite{hartmann2012weakly} formulate pixel-level segmentations as multiple instance learning weakly supervised  classifiers for a set of independent spatio-temporal segments. Tang~\emph{et al.}~\cite{tang2013discriminative} estimate the video in the positive sample with a large number of negative samples, and regard those segments with a distinct appearance as the foreground. Liu \emph{et al.}~\cite{liu2014weakly} further advance the study to address this problem in multi-class criterion rather than traditional binary classification. These methods rely on training examples and may produce inaccurate segmentation results. To overcome this limitation, Zhang~\emph{et al.}~\cite{zhang2015semantic} propose to segment semantic object in weakly labeled video by using object detection without the need of training process. In contrast, Tsai~\emph{et al.}~\cite{tsai2016semantic} does not require object proposal or video-level annotations. Their method link objects between different video and construct a graph for optimization.

Recently, Wang~\emph{et al.}~\cite{wang2016semi} combine the recognition and representation power of CNN with the intrinsic structure of unlabelled data in the target domain of weakly supervised semantic video object segmentation to improve inference performance. Unlike semantics-aware weakly-supervised methods, Khoreva~\emph{et al.}~\cite{khoreva2018video} employ natural language expressions to identify the target object in video. Their method integrate 
textual descriptions of interest as foreground into convnet-based techniques.

\vspace*{-2mm}

\subsection{Segmentation-based Tracking}
\label{sec:track}

\begin{figure}[t]
\centering
\begin{subfigure}{.7\textwidth}
\includegraphics[width=1\textwidth]{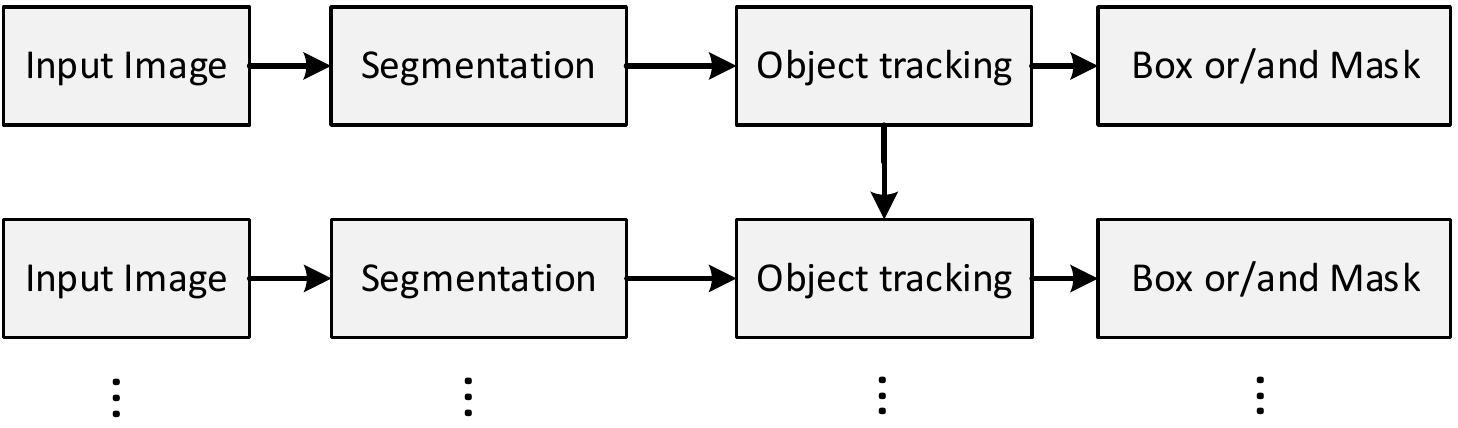} 
\caption{Bottom-up based framework.}
\label{fig:bottom-up-track}
\end{subfigure}
\begin{subfigure}{.7\textwidth}
\includegraphics[width=1\textwidth]{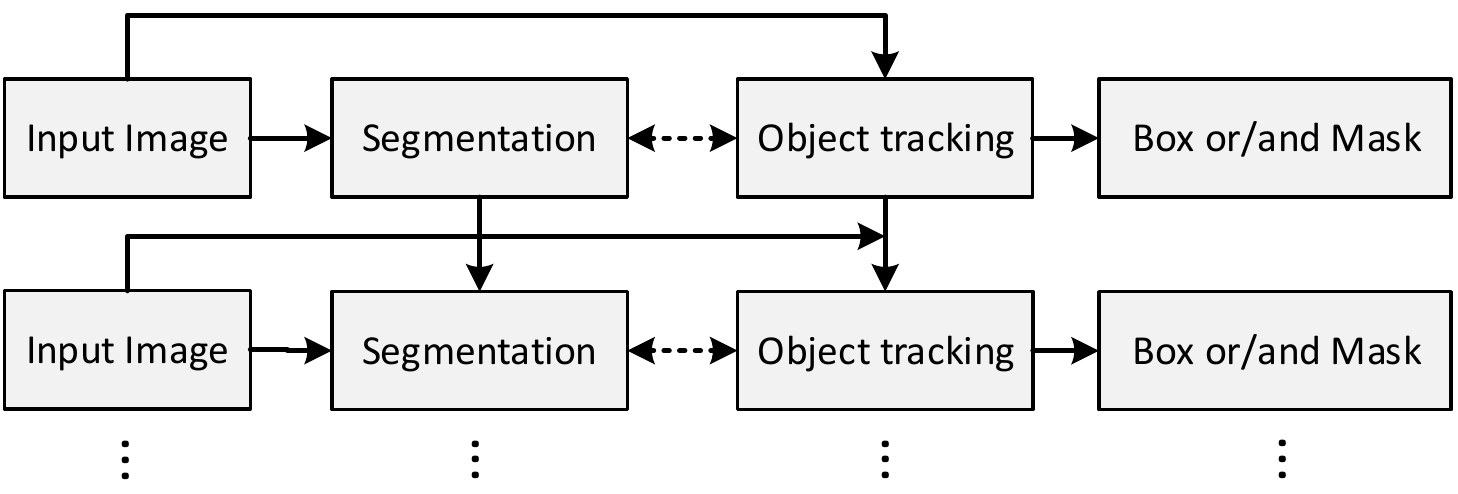} 
\caption{Joint-based framework.}
\label{fig:joint-track}
\end{subfigure}
\caption{Segmentation-based tracking frameworks: (a) Bottom-up based, (b) Joint-based.}
\label{fig:segment-based-track}
\end{figure}

In the previous video object segmentation methods, they usually cues like motion and appearance similarity to segment videos, that is, these methods estimate the position of a target in a manual or automatic manner. The object representation consists of a binary segmentation mask which indicates whether each pixel belongs to the target or not. For applications that require pixel-level information, such as video editing and video compression, this detailed representation is more desirable. Therefore, the estimating of all pixels requires a large amount of computational cost, and video object segmentation methods have been traditionally slow. In contrast, visual object tracking is to estimate the position of an object in the image plane as it moves around a scene. In general, the classical object shape is represented by a rectangle, ellipse, \emph{etc.} This simple object representation helps reduce the cost of data annotation. Moreover, such methods can quickly detect and track targets, and the initialization of object is relatively simple. However, these methods still operate more or less on the image regions described by the bounding box and are inherently difficult to track objects that undergo large deformations. To overcome this problem, some approaches integrate some form of segmentation into the tracking process. Segmentation-based tracking methods provide an accurate shape description for these objects. The strategies of these methods can be grouped into two main categories: bottom-up methods and joint-based methods. Figure~\ref{fig:segment-based-track} presents the flowchart of two segmentation-based tracking frameworks.

\subsubsection{Bottom-up based methods}
\label{sec:bottomup}

In the domain of bottom-up segmentation-based tracking, the object is presented from a segmented area instead of a bounding box. The segmentation-based tracking is a natural solution to handle non-rigid and deformable objects effectively. These methods use a low-level segmentation to extract regions in all frames, and then transitively match or propagate the similar regions across the video. We divide bottom-up based methods into two categories, namely contour matching and contour propagation. Contour matching approaches search for the object region in the current frame. On the other hand, by using a state space model,
contour propagation methods change the initial contour to a new position in the current frame.
A qualitative comparison of bottom-up segmentation-based tracking approaches is given in Tab.~\ref{tab:bottom-up}.

\paragraph{Contour matching} Contour matching searches the object silhouette and their associated models in the current frame. One solution is to build an appearance model of the object shape and match the best candidate image region to match the model, \emph{i.e.,} generative methods, for instance, integral histogram based models~\cite{1640835,5459276}, independent component analysis based models~\cite{Kong2010}, subspace learning based models~\cite{zhou2013shifted}, distance measures-based models~\cite{li2007automatic,colombari2007segmentation,hsiao2006robust}, spatio-temporal filter~\cite{889030}, spectral matching~\cite{cai2014robust}. Some approaches measure similarity between patches by comparing their gray-level or color histograms. Adam~\emph{et al.}~\cite{1640835} segment the target into a number of fragments to preserve the spatial relationships of the pixels, and use the integral histogram. Later, Chockalingam~\emph{et al.}~\cite{5459276} choose fragment adaptively according to the video frame, by clustering pixels with similar appearance, rather than using a fixed arrangement of rectangles. Yang~\emph{et al.}~\cite{Kong2010} propose a boosted color soft segmentation algorithm and incorporate independent component analysis with reference into the tracking framework. Zhou~\emph{et al.}~\cite{zhou2013shifted} present a shifted subspaces tracking to segment the motions and recover their trajectories. The authors use the Hausdorff distance~\cite{li2007automatic} and Mahalanobis distance~\cite{colombari2007segmentation} to construct a correlation surface from which the minimum is selected as the new object position. Hsiao~\emph{et al.}~\cite{hsiao2006robust} utilize trajectory estimation scheme for automatically deploying the growing seeds for tracking the object in further frames. Kompatsiaris~\emph{et al.}~\cite{889030} take into account the  intensity differences between consequent frames, and present a spatio-temporal filter to separate the moving person from the static background. Cai~\emph{et al.}~\cite{cai2014robust} build a dynamic graph to exploit the inner geometric structure information of the object by oversegmenting the target into several superpixels. And spectral clustering is used to solve the graph matching. 

\begin{table}[t]
\caption{Summary of bottom-up segmentation-based tracking methods. \#: number of objects, S: single, M: multiple. Box and Mask: the bounding box and mask of the object.}
\label{tab:bottom-up}
\centering
\scalebox{0.91}{
\begin{tabular}{p{2.2cm}|l|l|p{2.9cm}|p{2.8cm}|l}
\hline 
References & \# & Methods & Segment Techniques & Track Techniques & Results \\
\hline 
Son \cite{son2015tracking} & S & Contour matching & Graph-cut & Boosting decision tree & Box, Mask \\
\hline
Cai \cite{cai2014robust} & S & Contour matching & Graph-cut & SVM & Box \\
\hline
Duffner \cite{Duffner2014PixelTrack} & S & Contour matching & Probabilistic soft segmentation & Hough voting & Box \\
\hline
Wang \cite{Wang2012ROT} & S & Contour propagation & Mean shift clustering & Particle filter & Box \\
\hline
Zhou \cite{zhou2013shifted} & M & Contour matching & Subspace learning & Subspace learning & Mask \\
\hline
Heber \cite{HEBER2013573} & S & Contour matching & Graph-cut & Blending-based template, hough voting, mean-shift & Mask \\
\hline
Chien \cite{Chien2013Video} & M & Contour propagation & Threshold decision & Particle filter & Box \\
\hline 
Belagiannis \cite{belagiannis2012segmentation} & S & Contour propagation & Graph-cut & Particle filter & Mask \\
\hline
Godec \cite{godec2013hough} & S & Contour matching & Graph-cut & Hough voting & Mask \\
\hline
Wang \cite{wang2011superpixel} & S & Contour matching & SLIC & Particle filter & Box \\
\hline
Chockalingam \cite{chockalingam2009adaptive} & S & Contour matching & Spatially variant finite mixture models & Particle filter & Mask \\
\hline
Colombari \cite{colombari2007segmentation} & M & Contour matching & Region matching & Blob matching and connection & Mask \\
\hline
Hsiao \cite{hsiao2006robust} & S & Contour matching & Region growing and merging & Interframe difference & Mask \\
\hline
Kompatsiaris \cite{889030} & S & Contour matching & K-Means clustering & Spatiotemporal filter & Box, Mask \\
\hline
Gu \cite{718504} & S & Contour propagation & Morphological watershed & Temporal gradient & Mask \\
\hline
\end{tabular}
}
\end{table}

Another approach is to model both the object and the background and then to distinguish the object from the background by using a discriminative classifier, such as boosting-based models~\cite{son2015tracking}, Hough-based models~\cite{Duffner2014PixelTrack,godec2013hough}, and so on. These methods maintain object appearances based on small local patches or object regions, and perform tracking by classifying the silhouette into the foreground or the background. And the final tracking result is given by the mask of the best sample. Son~\emph{et al.}~\cite{son2015tracking} employ an online gradient decision boosting tree to classify each patch, and construct segmentation masks. Godec~\emph{et al.}~\cite{godec2013hough} propose a patch-based voting algorithm with Hough forests~\cite{gall2011hough}. By back-projecting the patches that voted for the object center, the authors initialize a graph-cut algorithm to segment foreground from background. However, the graph-cut segmentation it is relatively slow, and the binary segmentation increases the risk of drift due to wrongly segmented image regions. To address this problem, Duffner and Garcia~\cite{Duffner2014PixelTrack} present a fast tracking algorithm using a detector based on the generalized Hough transform and pixel-wise descriptors, then update the global segmentation model.

In addition, researchers propose hybrid generative-discriminative segmentation-based methods to fuse the useful information from the generative and the discriminative models. For instance, Heber~\emph{et al.}~\cite{HEBER2013573} present a segmentation-based tracking method to fuse three target tracker, \emph{i.e.} blending-based template tracker, Hough voting-based discriminative tracker, and feature histogram-based mean shift tracker. And the fusion process additionally provides a segmentation.

\paragraph{Contour propagation} Contour propagation of bottom-up based methods can be done using two different approaches: sequential Monte Carlo (or particle filter) based methods and direct minimization based methods. 
Some approaches employ sequential Monte Carlo-based methods to generate the state of the candidate of object contour~\cite{wang2011superpixel,Wang2012ROT,Chien2013Video,belagiannis2012segmentation}. The state is defined in terms of the shape and the motion parameters of the contour. Given all available observations of object $\mathcal{Z}_{1:t}=\{\mathcal{Z}_t, \dots, \mathcal{Z}_t\}$ up to the $t$-th frame, the state variable $\mathbf{y}_t$ is updated by the maximum a posteriori (MAP) estimation, \emph{i.e.} $\mathbf{y}_t=$argmax$_{\mathbf{y}_{t,i}}(\mathbf{y}_{t,i}|\mathcal{Z}_{1:t})$. 
The posterior probability $p(\mathbf{y}_{t,i}|\mathcal{Z}_{1:t})$ can be computed recursively as
\begin{align}
p(\mathbf{y}_t|\mathcal{Z}_{1:t}) \propto p(\mathcal{Z}_t|\mathbf{y}_t) \int p(\mathbf{y}_t|\mathbf{y}_{t-1})p(\mathbf{y}_{t-1}|\mathcal{Z}_{1:t-1})d\mathbf{y}_{t-1}.
\end{align} 
Here, $p(\mathcal{Z}_t|\mathbf{y}_t)$ is the observation model, which is usually defined in terms of the distance of the contour from observed edges. And $p(\mathbf{y}_t|\mathbf{y}_{t-1})$ represents the dynamic motion model. The dynamic motion model $p(\mathbf{y}_t|\mathbf{y}_{t-1})$ depicts the temporal correlation of state transition between two consecutive frames. The observation model $p(\mathcal{Z}_t|\mathbf{y}_t)$ describes the similarity between a candidate offset  and the best offset of the tracked object. For instance, Wang~\emph{et al.}~\cite{wang2011superpixel} use superpixel for appearance modeling and incorporate particle filtering to find the optimal target state, and their observation model is built as $p(\mathcal{Z}_t|\mathbf{y}_t) \propto C(\mathbf{y}_t)$, where $C(\mathbf{y}_t)$ represents the confidence of an observation at state $\mathbf{y}_t$. Belagiannis~\emph{et al.}~\cite{belagiannis2012segmentation} propose two particle sampling strategies based on segmentation to handle the object’s deformations, occlusions, orientation, scale and appearance changes. Some methods use particle-based approximate inference algorithm over the Dynamic Bayesian Network (DBN)~\cite{Wang2012ROT} and the Hidden Markov Model (HMM)~\cite{Chien2013Video} to estimate the contour. 


Both segmentation and tracking methods can minimize functions through gradient descent. In addition, Gu~\emph{et al.}~\cite{718504} combine supervised segmentation with unsupervised tracking. Specifically, the supervised segmentation method use mathematical morphology, and the unsupervised tracking method use computation of the partial derivatives. 


\subsubsection{Joint-based methods}
\label{sec:joint}

In the above bottom-up based methods, the foreground region is first segmented from the input image, then some features are extracted from the foreground region, and finally the object is tracked according to these features. The foreground segmentation and object tracking are performed as two separate tasks, as shown in Fig.~\ref{fig:segment-based-track} (a). The biggest limitation of these methods is that the errors in the foreground segmentation inevitably propagate forward, causing errors in object tracking. Therefore, many researchers integrate foreground segmentation and object tracking into a joint framework. The result of foreground segmentation determines the accuracy of feature extraction, which further affects the performance of silhouette tracking. On the other hand, the tracking results can provide top-down cues for foreground segmentation. These methods make full use of the correlation between foreground segmentation and object tracking, which greatly improve the performance of video segmentation and tracking, as shown in Fig.~\ref{fig:segment-based-track} (b). To utilize energy minimization techniques of the joint video object segmentation and tracking framework, we divide these methods into three categories, namely, graph-based framework, probabilistic framework, and CNN framework.  In Tab.~\ref{tab:joint-based}, we provide the qualitative comparison of these methods in this section.

\paragraph{Graph-based framework.} The basic technique of joint-based methods is to construct a graph for the energy function to be minimized. The variations on graph-based framework are primarily built using a small set of core algorithms-graph cuts~\cite{Bugeau2008,wen2015jots,8387770,8493320}, random walker~\cite{Papoutsakis2013}, and shortest geodesics~\cite{paragios2000geodesic}. 

\begin{table}[t]
\caption{Summary of joint segmentation-based tracking methods. \#: number of objects, S: single, M: multiple. Box and Mask: the bounding box and mask of the object.}
\label{tab:joint-based}
\centering
\scalebox{0.91}{
\begin{tabular}{p{2.0cm}|l|l|p{2.8cm}|p{3cm}|l}
\hline 
References & \# & Methods & Segment Technique & Track Technique & Results \\
\hline 
Keuper \cite{8493320} & M & Graph-based framework & Graph-cut & Deep matching & Box, Mask \\
\hline
Wang \cite{wang2018fast} & S & CNN & CNN & CNN & Box, Mask \\
\hline
Yao \cite{8387770} & S & Graph-based framework & CNN & Correlation filter & Box \\
\hline
Zhang \cite{zhang2018tracking} & S & CNN & CNN & CNN & Box, Mask \\
\hline
Yeo \cite{yeo2017superpixel} & S & Probabilistic framework & Markov Chain & Markov Chain & Box, Mask \\
\hline
Liu  \cite{liu2016visual} & M & Probabilistic framework & CRF & CRF & Mask \\
\hline
Tjaden \cite{tjaden2016real} & M & Probabilistic framework & Pixel-wise posterior & Pixel-wise posterior & Mask \\
\hline
Schubert \cite{schubert2015revisiting} & S & Probabilistic framework & Pixel-wise posterior & Pixel-wise posterior & Box, Mask \\
\hline
Milan \cite{milan2015joint} & M & Probabilistic framework & SVM & CRF & Box, Mask \\
\hline
Wen \cite{wen2015jots} & S & Graph-based framework & Graph-cuts & Energy minimization & Mask \\
\hline 
Papoutsakis \cite{Papoutsakis2013} & S & Graph-based framework & Random walker & Mean-shift & Box, Mask \\
\hline
Lim \cite{6751213} & S & Probabilistic framework & Graph-cuts & CRF & Mask \\
\hline
Aeschliman \cite{Aeschliman2010} & M & Probabilistic framework & Probabilistic soft segmentation & Probabilistic principal component analysis & Box, Mask \\
\hline
Wu \cite{Wu2009} & M & Probabilistic framework & Boosting & Boosting & Box \\
\hline 
Tao \cite{Tao2008Segmentation} & M & Probabilistic framework & MCMC & MCMC & Box \\
\hline
Bugeau \cite{Bugeau2008} & M & Graph-based framework & Graph-cuts & Mean-shift & Box, Mask \\
\hline
Bibby \cite{Bibby2008} & S & Probabilistic framework & Pixel-wise posterior & Pixel-wise posterior & Box, Mask \\
\hline
Paragios \cite{paragios2000geodesic} & M & Graph-based framework & Active contours & Interframe difference & Mask \\
\hline
\end{tabular}
}
\end{table}

For instance, Bugeau and P{\'{e}}rez~\cite{Bugeau2008} formulate an objective functions that combine low-level pixel-wise measures and high-level observations. The minimization of these cost functions simultaneously allows tracking and segmentation of tracked objects. In~\cite{wen2015jots}, Wen~\emph{et al.} integrate the multi-part tracking and segmentation into a unified energy minimization framework, which is optimized iteratively by a RANSAC-style approach. Yao~\emph{et al.}~\cite{8387770} present a joint framework to introduce semantics~\cite{lin2017refinenet} into tracking procedure. Then, they propose to exploit semantics to localise object accurately via an energy-minimization-based segmentation. In~\cite{8493320}, Keuper~\emph{et al.} present a graph-based segmentation and multiple object tracking framework. Specifically, they combine bottom-up motion segmentation by grouping of point trajectories with top-down multiple object tracking by  clustering of bounding boxes. The random walker algorithm~\cite{Papoutsakis2013} is also formulated on a weighted graph. The joint framework integrates the EM-based object tracking and Random Walker-based image segmentation in a closed loop scheme. In addition, Paragios and Deriche~\cite{paragios2000geodesic} present a graph-based framework to link the minimization of a geodesic active contour objective function to the detection and the tracking of moving objects.

\paragraph{Probabilistic framework.} There are many probabilistic framework for jointly solving video object segmentation and tracking, such as Bayesian methods~\cite{Aeschliman2010,Tao2008Segmentation,Wu2009,yeo2017superpixel}, pixel-wise posterior based methods~\cite{Bibby2008,schubert2015revisiting,tjaden2016real}, and CRF based methods~\cite{liu2016visual,6751213,milan2015joint}. In~\cite{Aeschliman2010}, Aeschliman~\emph{et al.} present a probabilistic framework for jointly solving tracking and fine, pixel-level segmentation. The candidate target locations are evaluated by first computing a pixel-level segmentation, and explicitly including this segmentation in the probability model. Then  the segmentation is used to incrementally update the probability model. In addition, Zhao~\emph{et al.}~\cite{Tao2008Segmentation} propose a Bayesian framework that integrates segmentation and tracking based on a joint likelihood for the appearance of multiple objects, and perform the inference by an Markov chain Monte Carlo-based approach. Later, Wu and Nevatia~\cite{Wu2009} present a joint framework to take the detection results as input and search for the multiple object configuration with the best image likelihood. Yeo~\emph{et al.}~\cite{yeo2017superpixel} employ absorbing Markov Chain algorithm over superpixel segmentation to estimate the object state, and target segmentation is propagated to subsequent frames in an online manner.

In~\cite{Bibby2008,schubert2015revisiting,tjaden2016real}, a probability generative model is built a segmentation-based tracking method using pixel-wise posteriors. These methods construct the appearance-model using a probabilistic formulation, carry out the level-set segmentation using this model, and then perform the contour propagation. The minimization of these algorithms are implemented by the gradient descent. Thereinto, Tjaden~\emph{et al.}~\cite{tjaden2016real} segment multiple 3D objects and track pose using pixel-wise second-order optimization approach.

Some methods utilize energy minimization techniques of CRF to perform fine segmentation and target object. For instance, Milan~\emph{et al.}~\cite{milan2015joint} propose a CRF model that exploits high-level detector responses and low-level superpixel information to jointly track and segment multiple objects. Lim~\emph{et al.}~\cite{6751213} handle joint estimation to segment foreground object and track human pose using a MAP solution. Liu~\emph{et al.}~\cite{liu2016visual} present a unified dynamic couple CRF model to joint track and segment moving objects in region level.

\paragraph{CNN framework.} Recently, some researchers begin to pay attention to perform visual object tracking and semi-supervised video object segmentation using convolutional neural network framework~\cite{wang2018fast,zhang2018tracking}. Wang~\emph{et al.}~\cite{wang2018fast} present a a Siamese network to simultaneously estimate binary segmentation mask, bounding box, and the corresponding object/background scores. By only inputting a bounding box in the first frame, Zhang~\emph{et al.}~\cite{zhang2018tracking} build a two-branch network, \emph{i.e.,} appearance network and contour network. And tracking output and segmentation results help refine each other mutually.


\subsubsection{Discussion.} In general, given a bounding box of a target in the first frame, the bottom-up based methods estimate the location of the object in subsequent images, which is similar to the tracking-by-detection methods. Unlike traditional visual object tracking methods, the bottom-up based methods use the foreground object contour as a special feature to solve the problem of object drift in non-rigid object tracking and segmentation. The purpose of these methods is to find the location of the target, so only the bounding box or coarse mask of object is estimated. Some methods simply use the result of the segmentation to estimate the scale problem in visual object tracking. Compared to joint-based methods, the processing speed of these methods is faster.

On the other hand, joint-based methods unify the two tasks of segmentation and tracking into the graph-based or probabilistic framework, and use energy minimization method to estimate the exact object mask. Specifically, these energy minimization methods are iterated many times to estimate accurate object poses, motions, occlusions, and so on. Many methods do not output the bounding box of the object, but only the object mask. Generally, iterative optimization inherently limits runtime speed. Recently, some researchers have used offline and online CNN-based methods to simultaneously process segmentation and tracking, and the impressive results demonstrate accurate and very fast tracking and segmentation.


\section{Datasets and metrics} 
\label{sec:dataset_metric}

To evaluate the performance of various video object segmentation and tracking methods, one needs test video dataset, the ground truth, and metrics of the competing approaches. In this section, we will give brief introduced of datasets, evaluation protocols.

\subsection{Video object segmentation and tracking datasets}
\label{sec:dataset}

A brief summary of video object segmentation and tracking datasets is shown in Table~\ref{tab:dataset} . These datasets are described and discussed in more detail next.

\begin{table}[t]
\caption{Brief illustration of datasets that are used in the evaluation of the video object segmentation and tracking methods. V \#: number of video. C \#: number of categories. O \#: number of objects. A \#: annotated frames. U, S, I, W, T: unsupervised VOS, semi-supervised VOS, interactive VOS, weakly supervised VOS, and segmentation-based tracking methods. Object pro.: object property, T. of methods: type of methods. }
\label{tab:dataset}
\centering
\scalebox{.9}{
\begin{tabular}{l|c|c|c|c|c|c|c|l}
\hline 
 & V \# & C \# & O \# & A \# & Object pro. & T. of methods & Labels & Publish year \\
\hline
\hline
\emph{SegTrack} & 6 & 6 & 6 & 244 & Single & U, S, I, W, T & Mask & 2012~\cite{Tsai2012} \\ 
\hline
\emph{SegTrack v2} & 14 & 11 & 24 & 1475 & Multiple & U, S, I, W, T & Mask & 2014~\cite{Li2014Video} \\
\hline
\emph{BMS-26} & 26 & 2 & 38 & 189 & Multiple & U & Mask & 2010~\cite{Brox2010} \\
\hline
\emph{FBMS-59} & 59 & 16 & 139 & 1,465 & Multiple & U, S & Mask & 2014~\cite{Peter2014Segmentation} \\
\hline
\emph{YouTube-objects} & 126 & 10 & 96 & 2,153 & Single & U, S, I, W & Mask & 2014~\cite{prest2012learning,Jain243} \\
\hline
\emph{YouTube-VOS} & 3252 & 78 & 6048 & 133,886 & Multiple & S & Mask & 2018~\cite{Xu_2018_ECCV}\\ 
\hline
\emph{JumpCut} & 22 & 14 & 22 & 6,331 & Single & U, S & Mask & 2015~\cite{fan2015jumpcut} \\
\hline
\emph{DAVIS 2016} & 50 & -- & 50 & 3,440 & Single & U, S, I, W & Mask & 2016~\cite{perazzi2016benchmark} \\
\hline
\emph{DAVIS 2017} & 150 & -- & 384 & 10,474 & Multiple & U, S, I, W & Mask & 2017~\cite{pont20172017} \\
\hline
\emph{NR} & 11 & -- & 11 & 1,200 & Single & S, T & Box, Mask & 2015~\cite{son2015tracking} \\
\hline
\emph{MOT 2016} & 14 & -- & -- & 11,000 & Multiple & U, T & Box & 2016~\cite{milan2016mot16} \\
\hline
\emph{VOT 2016} & 60  & -- & 60 & 21,511 & Single & S, T & Box, Mask & 2016~\cite{10.1007_978-3-319-48881-3_54,vojir2017pixel} \\
\hline
\emph{OTB 2013} & 50 & -- & 50 & 29,000 & Single & T & Box & 2013~\cite{wu2013online} \\
\hline
\emph{OTB 2015} & 100 & -- & 100 & 58,000 & Single & T & Box & 2015~\cite{Wu2015Object} \\
\hline
\end{tabular}
}
\end{table}

{\bf SegTrack}~\cite{Tsai2012} and {\bf SegTrack v2}~\cite{Li2014Video} are introduced to evaluate tracking and video object segmentation algorithms. SegTrack contains 6 videos (\emph{monkeydog, girl, birdfall, parachute, cheetah, penguin}) and pixel-level ground-truth for the single moving foreground object in every frame. These videos provide a variety of challenges, including non-rigid deformation, similar objects and fast motion of the camera and target. SegTrack v2 contains 14 videos with instance-level moving object annotations in all the frames. Other videos from SegTrack v2 also include cluttered backgrounds and dynamic scenes caused by camera movement or moving background objects. In some video sequences, the objects are visually very similar to the image background, that is, low contrast along object boundaries, such as the \emph{birdfall, frog} and \emph{worm} sequences in the SegTrack v2 dataset. In contrast to SegTrack, many videos have more than one object of interest in SegTrack v2.

{\bf BMS-26} (Berkeley motion segmentation)~\cite{Brox2010} and {\bf FBMS-59} (Freiburg-Berkeley motion segmentation)~\cite{Peter2014Segmentation} are widely used for unsupervised and semi-supervised VOS methods. BMS-26 dataset consists 26 videos with a total of 189 annotated image frames, which shots from movie stories and the 10 vehicles and 2 human sequences. The FBMS-59 dataset reflects two major improvements in the previous version of BMS-26. First, the updated version dataset adds 33 new sequences, therefore, the FBMS-59 dataset consists of 59 sequences. Second, these 33 new sequences incorporate challenges of unconstrained videos such as fast motion, motion blur, occlusions, and object appearance changes. The sequences are divided into 29 training and 30 test video sequences.


{\bf YouTube-objects}~\cite{prest2012learning} and {\bf YouTube-VOS} (YouTube video object segmentation)~\cite{Xu_2018_ECCV} contain a large amount of Internet videos. Jain~\emph{et al.}~\cite{Jain243} adopt its subset that contains 126 videos with 10 object categories and 20,977 frames. In these videos, 2,153 key-frames are sparsely sampled and manually annotated in pixel-wise masks according to the video labels. YouTube-objects dataset is used for unsupervised, semi-supervised, interactive, and weakly supervised VOS approaches. In 2018, Xu~\emph{et al.}~\cite{Xu_2018_ECCV} release a large-scale video object segmentation dataset called YouTube-VOS. The dataset contains 3,252 YouTube video clips and 133,886 object annotations, of which 78 categories cover  78 categories covering common animals, cars, accessories and human activities. At the same time, the authors build a sequence-to-sequence semi-supervised video object segmentation algorithm to verify this dataset and performance. 

{\bf JumpCut} dataset~\cite{fan2015jumpcut} consists of 22 video sequences with medium image resolution. It contains 14 categories (6,331 annotation images in total) along with pixel level ground-truth annotations. Most image frames in the JumpCut dataset contain very fast object motion and significant foreground deformations. Thus, the JumpCut dataset is considered a more challenging video sequences for unsupervised and semi-supervised VOS, and is widely used to evaluate modern unsupervised and semi-supervised video segmentation techniques.

{\bf DAVIS 2016, 2017, and 2018} datasets are one of the most popular datasets for training and evaluating video object segmentation algorithms. DAVIS 2016~\cite{perazzi2016benchmark} dataset contains 50 full high quality video sequences with 3,455 annotated frames in total, and focuses on single-object video object segmentation, that is, there is only one foreground object per video. 30 training set and 20 validation set in this dataset is divided. Later, DAVIS 2017~\cite{pont20172017} complements DAVIS 2016 dataset training and validation sets with 30 and 10 high quality videos, respectively. It also provides an additional 30 development test sequences and 30 challenge test sequences. Also, the DAVIS 2017 dataset relabels multiple objects in all video sequences. These improvements make it more challenging than the original DAVIS 2016 dataset. In addition, Each video is labeled with multiple attributes such as occlusion, object deformation, fast motion, and scale change to provide a comprehensive analysis of model performance. Moreover, DAVIS 2018 dataset~\cite{caelles20182018} adds 100 videos with multiple objects per video to the original DAVIS 2016 dataset, and complements an interactive segmentation teaser track.

{\bf NR} (non-rigid object tracking) dataset~\cite{son2015tracking} consists of 11 video sequences with 1,200 frames which contain deformable and articulated objects. First, the pixel-level annotations are performed manually. The bounding box annotation is then generated by calculating the tightest rectangular bounding box that contains all of the object pixels. Each video is labeled with only one object. It has been to evaluate segmentation-based tracking and semi-supervised VOS algorithms.

{\bf MOT 2016} (multiple object tracking) dataset~\cite{milan2016mot16} consists of 14 sequences with 11,000 frames which contain crowded scenarios, different viewpoints, camera and object motions and weather conditions. The targets are annotated with axis-aligned minimum bounding boxes in each video sequence.
The scale of datasets for MOT is relatively smaller than single object tracking, and current datasets focus on pedestrians. This dataset is used to evaluate multiple object of unsupervised VOS and segmentation-based tracking algorithms.

{\bf VOT 2016} (video object tracking) dataset~\cite{10.1007_978-3-319-48881-3_54} contains 60 high-quality video sequences targeted at single video tracking and segmentation tasks. It consists 21,511 frames in total. In~\cite{vojir2017pixel}, Vojir~\emph{et al.} provide pixel-level segmentation annotations for the VOT 2016 dataset, and construct a challenging segmentation tracking and test dataset.

{\bf OTB} (object tracking benchmark) is widely used to evaluate single segmentation-based video object tracking algorithms. OTB 2013
dataset~\cite{wu2013online} has 50 video sequences includes fully annotated video sequences with bounding box. OTB 2015 dataset~\cite{Wu2015Object} consists of 100 video sequences and 58,000 annotated frames of real-world moving objects.

\subsection{Metrics}
\label{sec:metric}

To order to evaluate the performance, this section focuses on the specific case of video object segmentation and tracking, where both the predicted results and the ground-truth are used for foreground-background partitioning. Measures can be focused on evaluating which pixels of the ground truth are detected, or indicating the precision of the bounding box.

\subsubsection{Evaluating of pixel-wise object segmentation techniques}
\label{sec:metric_pixel}
For video object segmentation, the standard evaluation metric has three measurements~\cite{perazzi2016benchmark}, namely the spatial precision of the segmentation, the consistency of contour similarity and the temporal stability. 
\begin{itemize}
\item Region similarity $\mathcal{J}$. The region similarity is the \emph{intersection over union} (IoU) function between the predicted object segmentation mask $M$ and ground truth $G$. This quantitative metric for measuring the number of misclassified image pixels and measuring pixels matching segmentation algorithm. In this way, it is defined as $\mathcal{J}=\frac{M\cap G}{M\cup G}$.
\item Contour precision $\mathcal{F}$. The segmented mask is treated as a set of closed contour regions, and the function of precision and recall is to calculate the contour-based $F$-measure. That is to say, the $F$-measure of the contour precision is based on the precision and recall of the contour. This indicator is used to measure the precision of the segmentation boundary. Let segmented mask $M$ be interpreted as a set of closed contours $c(M)$. Thus, we can achieve the contour-based precision $P_c$ and recall $R_c$ based on $c(M)$ and $c(G)$. Therefore, $F$-measure is defined as $\mathcal{F}=\frac{2P_c R_c}{P_c+R_c}$.
\item Temporal stability $\mathcal{T}$. Most of VOS methods also use time stability $t$ to measure the turbulence and inaccuracy of the contours. The temporal stability of the video segmentation is measured by the dissimilarity of the target shape context descriptors that describe the pixels on the contour of the segmentation between two adjacent frames in the video sequences.
\end{itemize}

\subsubsection{Evaluating of bounding box based object tracking techniques}
\label{sec:metric_box}
For segmentation-based tracking approaches, both the mask and the bounding box of object may be output. To evaluate object tracking algorithms, therefore, we should account for two categories: single object and multiple objects. 

The evaluation protocol of OTB 2013~\cite{wu2013online} and VOT 2016~\cite{10.1007_978-3-319-48881-3_54} dataset is widely used in single object tracking algorithm. For OTB 2013 benchmark, four metrics with one-pass evaluation (OPE) are used to evaluate all the compared trackers: (i) bounding box overlap, which is measured by VOC overlap ratio (VOR); (ii) center location error (CLE), (iii) distance precision (DP), and (iv) overlap precision (OP). For VOT 2016 benchmark, there are three main measures for analyzing the performance of short-term tracking: accuracy, robustness, and expected average overlap (EAO). The accuracy is the average overlap between the prediction during the successful tracking and the real boundary box of the ground truth. The robustness measures the number of times a tracker loses a target (\emph{i.e.}, fails) during the tracking period. EAO estimates the accuracy of the estimated bounding box after processing a certain number of frames since initialization.

Metrics for multiple targets tracking are divided into three classes by different attributes: accuracy, precision, and completeness. Combining multiple target false positives, false positives, and mismatches into a single value becomes a multi-target tracking accuracy (MOTA) metric. The multiple object tracking accuracy (MOTP) metric describes the accuracy of measuring objects by boundary box overlap and/or center location distance. The complete metrics indicate the completeness of tracking the ground truth trajectory.

\section{Future directions}
\label{sec:future}

Based on significant advances in video object segmentation and tracking, we suggest some future research directions that would be interesting to pursue.

\noindent{\bf Simultaneous prediction of VOS and VOT.} In the traditional hand-crafted video object segmentation and tracking methods, there are many algorithms to simultaneously output the mask and bounding box of the object. Recently, researchers came up with end-to-end VOS and VOT methods that dealt with two problems in a deep framework, they simultaneously predict pixel-level object masks and object-level bounding boxes for impressive performance.
This will lead to an important problem: speed and accuracy. On the one hand, accuracy is important in some applications, such as fine-tuning iterations to improve segmentation and tracking performance. It is computationally expensive and the speed is bound to be slow. On the other hand, if the processing speed is increased without losing the performance of object segmentation and tracking, this will be a very interesting direction.

\noindent{\bf Fine-grained video object segmentation and tracking.} Segmentation and tracking of fine-grained objects in the full HD video is challenging. Since such videos generally have a large background of various appearance and motion, small parts of fine-grained objects in video cannot be segmented and tracked with sufficient accuracy. On the one hand, in fine-grained segmentation and recognition tasks, these small parts usually contain semantic information that is extremely important for fine-grained classification. Moreover, object tracking is essentially a process of continuous predicting the motion of a very small object between frames, if a method can not distinguish good small differences, it may not be an optimal design choice. Therefore, how to accurately segment and track fine-grained objects, and then improve the performance of video and video recognition tasks plays an important role in many real-world applications. 

\noindent{\bf Generalization performance of VOST.} Generalization has always been a difficulty in video segmentation and tracking algorithms. Although VOST tasks can be solved after training, but it is difficult to transfer the acquisition experience to new categories, or unconstrained videos, such as these videos are noisy, compressed, unstructured, and included by moving from multiple views. End-to-end training in deep learning is currently used to improve generalization. Although there are many datasets, such as DAVIS, YouTube-VOS, OTB, and VOT, these datasets have some limitations and are somewhat different from the actual environment. Not only the diversity of the appearance of the foreground object, but also the complexity of the object motion trajectory will directly affect the generalization ability of the object segmentation and tracking methods. Therefore, how to fast and accurate segment and track objects in these new categories or environments will be the focus of research.

\noindent{\bf Multi-camera video object segmentation and tracking.} Performing video analysis and monitoring in complex environments requires the use of multiple cameras. This problem has led to an increasing interest in research on multi-camera collaborative video analysis. In multiple cameras, due to the fusion of different visual information from different viewpoints, the method synergistically handles video object segmentation and tracking in the same scene monitored by different cameras, thus it may improve the performance. However, it should be noted that the images used in multi-camera surveillance are usually captured by cameras located at different locations. Therefore, there is a great diversity in visual perspective, which should be considered separately in the video object segmentation and tracking techniques.

\noindent{\bf 3D video object segmentation and tracking.} The analysis and processing of 3D object is a core problem in the computer vision community. There may be two directions of interest here. First, VOST is an important prerequisite for avoiding obstacles and pedestrians. Segmentation combined with 3D images produces detailed object boundaries in 3D. The subsequent path planning algorithm can then generate motion trajectories to avoid collisions. Autonomous robots use video object segmentation and tracking to locate, find and grab objects of interest. Second, in building infrastructure modeling, you can create virtual 3D models of buildings that contain their semantic regions. This model can then be used to quickly calculate statistics in the video. The 3D reconstruction system provides very detailed geometry. However, after scanning, cumbersome post-processing steps are required to cut the object of interest. Video object segmentation and tracking helps automate this task. 

\section{Conclusion}
\label{sec:conclusion}

In this article, we provided a comprehensive survey of the video object segmentation and tracking literature. We described challenges and potential application in the field, classified and analyzed the recent methods, and discussed different algorithms. The presented survey uses an organization of application scenarios to review five important categories of literature in VOST: unsupervised VOS, semi-supervised VOS, interactive VOS, weakly supervised VOS, and segmentation-based tracking methods. We provided a hierarchical categorization of the different groups in existing works, and summarized some object representation, image features, motion cues, \emph{etc}. We also described various of per-process and post-process CNN-based VOS methods, and discussed the advantages or disadvantages aspects of the methods. Moreover, we described the related video datasets for video object segmentation and tracking, and the evaluation metrics of pixel-wise mask and bounding box based techniques. We believe this review will benefit researchers in this field and provide useful insights into this important research topic. We hope to encourage more future work to develop in this direction.






\end{document}